%% file: egpaper_final.tex
\documentclass[10pt,twocolumn,letterpaper]{article}

\usepackage{iccv}
\usepackage{times}
\usepackage{epsfig}
\usepackage{graphicx}
\usepackage{amsmath}
\usepackage{amssymb}
\usepackage{subcaption}
\usepackage{blindtext}
\usepackage{booktabs}  
\usepackage{multirow}
\usepackage{enumitem}
\usepackage{pifont}
\usepackage{tabulary,multirow,overpic,xcolor,subfloat}
\usepackage{colortbl}
\usepackage[pagebackref=true, breaklinks=true, letterpaper=true, colorlinks, bookmarks=false]{hyperref}
\usepackage{cleveref}
\usepackage{algorithm}
\usepackage{algorithmic}
\usepackage{listings}

\definecolor{linkcolor}{HTML}{ED1C24}
\definecolor{graycolor}{rgb}{0.95,0.95,0.95}

\newcommand{\figref}[1]{Fig.~\ref{#1}}
\newcommand{\tabref}[1]{Tab.~\ref{#1}}
\newcommand{\xmark}{\ding{55}}
\newcommand{\cmark}{\ding{51}}
\newcommand{\R}{\mathbb{R}}

\newcommand\blfootnote[1]{%
  \begingroup
  \renewcommand\thefootnote{}\footnote{#1}%
  \addtocounter{footnote}{-1}%
  \endgroup
}

 \iccvfinalcopy %

\def\iccvPaperID{****} %

\ificcvfinal\fi

\begin{document}

\title{Hybrid Spectral Denoising Transformer with Guided Attention}

\author{Zeqiang Lai$^1$, \quad  Chenggang Yan$^2$, \quad Ying Fu$^{1\dagger}$\\
$^1$Beijing Institute of Technology \quad $^2$ Hangzhou Dianzi University\\
{\tt\small \{laizeqiang, fuying\}@bit.edu.cn \quad cgyan@hdu.edu.cn}
}

\maketitle
\ificcvfinal\thispagestyle{empty}\fi

\blfootnote{$\dagger$ Corresponding Author.}

\input{section/abstract.tex}
\input{section/introduction.tex}
\input{section/related_work.tex}

\input{section/method.tex}

\input{section/experiments.tex}

\input{section/conclusion.tex}

\section*{Acknowledgement}

We thank many colleagues for their help, in particular, Zhiyuan Liang for the paper revision. Miaoyu Li for useful discussions; Wenzhuo Liu for help on computational resources.
This work was also generously supported by the National Natural Science Foundation of China under Grants No. 62171038, No. 61827901, and No. 62088101.

{\small
\bibliographystyle{ieee_fullname}
\bibliography{egbib}
}

\clearpage
\input{section/appendix.tex}

\end{document}

%% file: section/abstract.tex
\begin{abstract}

 In this paper, we present a Hybrid Spectral Denoising Transformer (HSDT) for hyperspectral image denoising. Challenges in adapting transformer for HSI arise from the capabilities to tackle existing limitations of CNN-based methods in capturing the global and local spatial-spectral correlations while maintaining efficiency and flexibility. 
 To address these issues, we introduce a hybrid approach that combines the advantages of both models with a Spatial-Spectral Separable Convolution (S3Conv), Guided Spectral Self-Attention (GSSA), and Self-Modulated Feed-Forward Network (SM-FFN). 
 Our S3Conv works as a lightweight alternative to 3D convolution, which extracts more spatial-spectral correlated features while keeping the flexibility to tackle HSIs with an arbitrary number of bands. 
 These features are then adaptively processed by GSSA which performs 3D self-attention across the spectral bands, guided by a set of learnable queries that encode the spectral signatures. This not only enriches our model with powerful capabilities for identifying global spectral correlations but also maintains linear complexity. 
 Moreover, our SM-FFN proposes the self-modulation that intensifies the activations of more informative regions, which further strengthens the aggregated features. 
 Extensive experiments are conducted on various datasets under both simulated and real-world noise, and it shows that our HSDT significantly outperforms the existing state-of-the-art methods while maintaining low computational overhead. 
 Code is at \url{https://github.com/Zeqiang-Lai/HSDT}.

\end{abstract}

%% file: section/introduction.tex
\section{Introduction}

Hyperspectral image (HSI) provides substantially more abundant spectral information than the ordinary color image, which makes it especially utilitarian in the field of remote sensing \cite{bioucas2013hyperspectral, blackburn2007hyperspectral}, biometric authentication \cite{uzair2015hyperspectral}, detection \cite{mehta2021dark}, and geological science \cite{ellis2001searching,smith2006introduction}. Nevertheless, limited by imaging techniques, most existing HSI cameras still suffer from various types of noise that might degrade the performance of their applications, which urges the development of robust HSI denoising algorithms. 

\begin{figure}[t]
\centering
\includegraphics[width=1\linewidth]{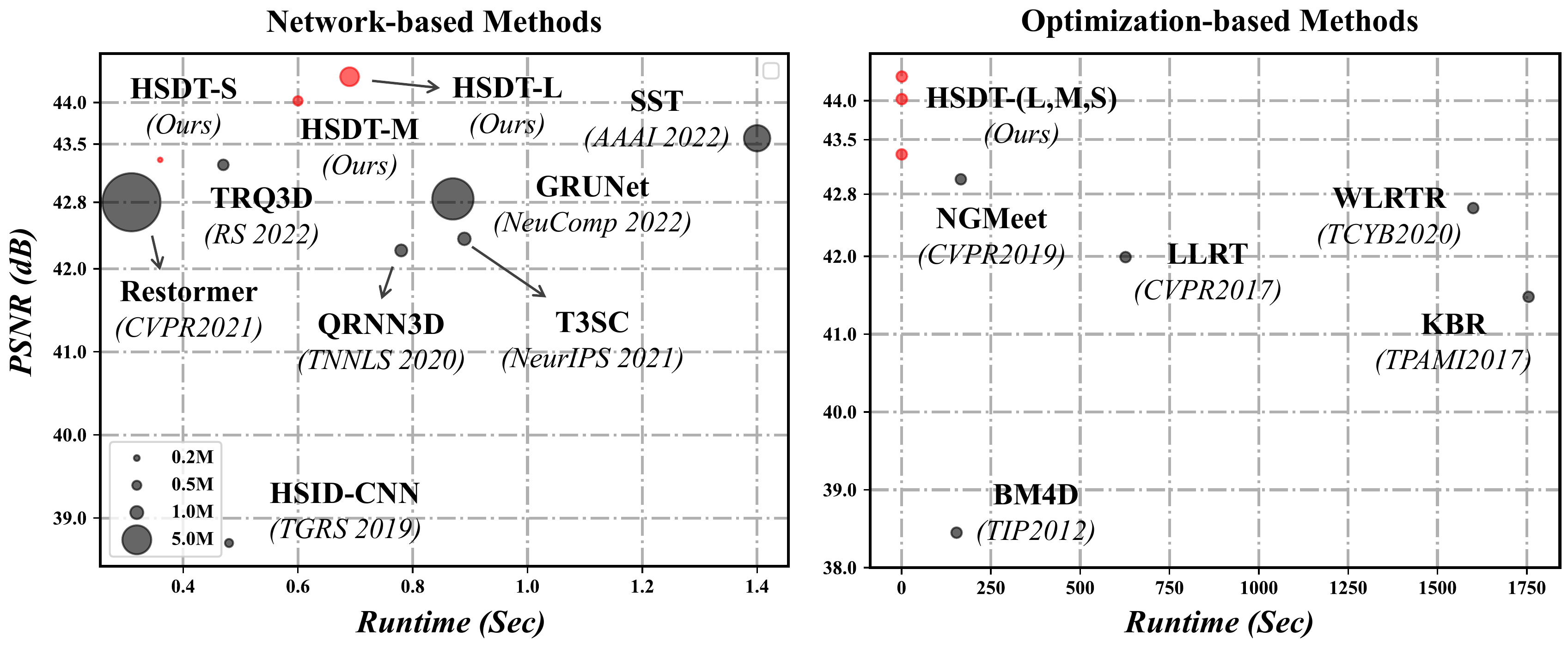}
\vspace{-6mm}
\caption{Our method achieves state-of-the-art performance while maintaining low computational overhead.}
\label{fig:flops}
\vspace{-2mm}
\end{figure}

Motivated by the intrinsic properties of HSI, traditional HSI denoising approaches \cite{yuan2012hyperspectral, he2019non} often exploit the optimization schemes with priors, \eg, low rankness \cite{zhao2020fast,wei2019low}, non-local similarities \cite{maggioni2012nonlocal,mei2021image}, spatial-spectral correlation \cite{peng2014decomposable}, and global correlation along the spectrum \cite{wei20203}.
Whilst offering appreciable performance, the efficacy of these methods is largely dependent on the degree of similarity between the handcrafted priors and the real-world noise model, and these methods are often challenging to accelerate with modern hardwares due to the complex processing pipelines.
Recent HSI denoising methods based on Convolutional Neural Network (CNN) \cite{yuan2018hyperspectral, lai2022deep, bodrito2021trainable} get rid of handcrafted regularizations with learning-based prior and often run faster with graphic accelerators and machine-learning frameworks \cite{paszke2019pytorch}. However, these methods are still insufficient for exploring the characteristics of HSI, \eg, {global and local spectral-spatial correlations}. For example, HSID-CNN \cite{yuan2018hyperspectral} only considers the correlations between several adjacent spectral bands. QRNN3D \cite{wei20203} and GRUNet \cite{lai2022deep} model the global spectral correlations with quasi-recurrent units \cite{bradbury2016quasi} but suffer from the problem of vanished correlations for long-range separate bands due to the recurrent multiplications of merging weights. 
Besides, recent methods \cite{wei20203, zhang2021hyperspectral} tend to use 3D convolution to explore the local spectral-spatial correlations while maintaining the flexibility to handle different HSIs. 
This strategy, however, introduces substantially unwanted computation and parameters.

Starting from natural language processing \cite{vaswani2017attention}, transformer architectures \cite{dosovitskiy2020image,bao2021beit} have recently been applied to various vision tasks including color image restoration \cite{liang2021swinir,zamir2022restormer} and HSI processing \cite{li2022spatial,cai2022mask,pang2022trq3dnet}. With the multi-head self-attention of transformer, these methods enjoy stronger capabilities of capturing non-local similarity and long-range dependency over aforementioned CNN-based methods. Despite of that, they are still suboptimal and inflexible for diverse HSIs. 
On the one hand, existing attentions for HSIs apply along either spatial \cite{li2022spatial,pang2022trq3dnet} or 2D feature channel \cite{li2022spatial,cai2022mask} dimensions, which could introduce quadratic complexities or break down the structured spectral dependency. 
On the other hand, their 2D architectural designs also make their models specifically bound to one type of HSI, \eg, HSI with 31 bands, and separate models have to be trained for other types, \eg, HSI with 210 bands. This can be problematic since the amount of available datasets is unevenly distributed for different HSIs.  
Finally, HSIs often exhibit beneficial fixed structures, \eg, relative intensity correlations of different bands for objects. Direct transfer of existing transformer blocks without considering this fact might lead to suboptimal performance for HSI denoising.

\begin{figure}[t]
\centering
\includegraphics[width=1\linewidth]{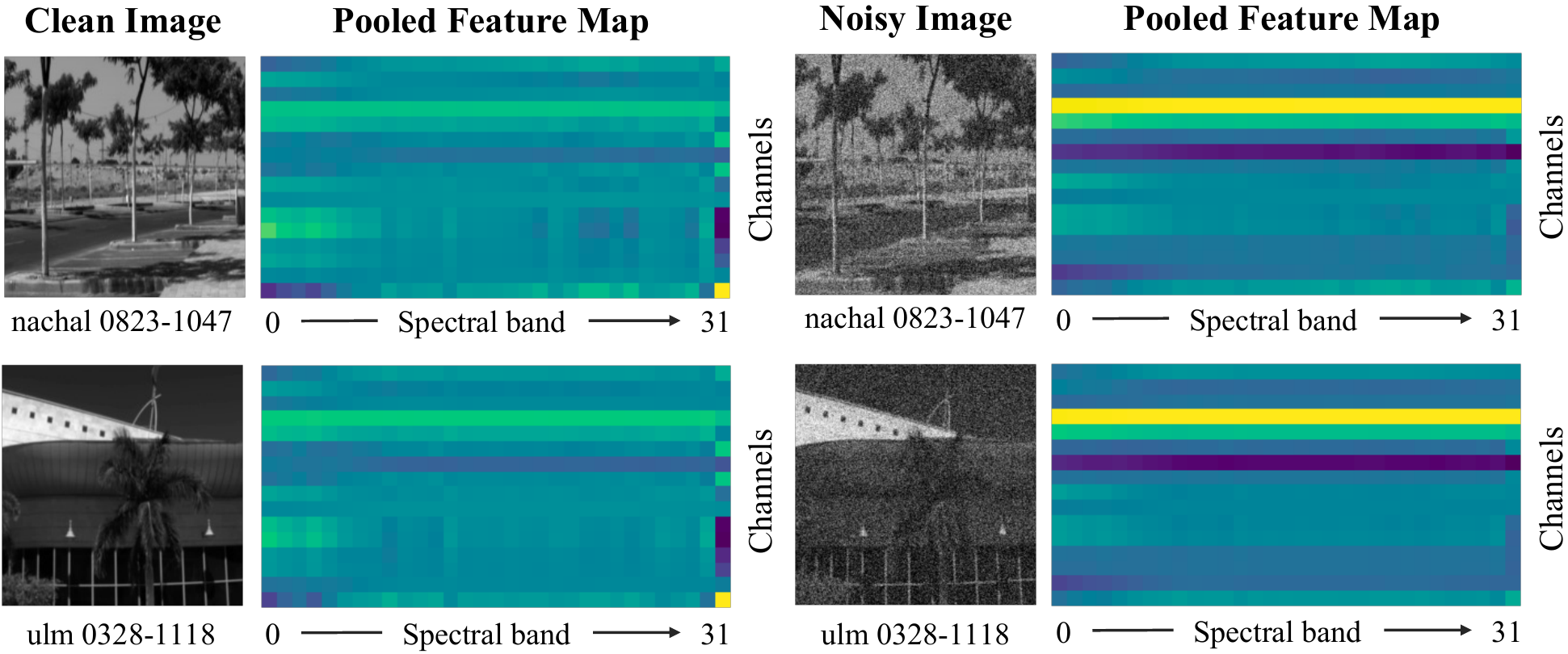}
   \caption{Visualization of feature maps of different bands by performing global average pooling on spatial locations. It can be observed that despite the difference of images, the pooled feature maps of clean and noisy images share different common patterns, which might be helpful for identifying, \eg, cleaner bands, for denoising noisy bands.}
\label{fig:stat}
\vspace{-2mm}
\end{figure}

In this paper, we propose a novel Hybrid Spectral Denoising Transformer (HSDT) that effectively integrates the local spectral-spatial inductive bias of the convolution and the long-range spectral dependency modeling ability of the transformer. Unlike previous HSI transformers \cite{li2022spatial, cai2022mask, pang2022trq3dnet}, HSDT is designed to be effective and flexible for handling diverse HSIs in a single model, which results in a variety of benefits, \eg, the ability to jointly utilize HSIs with different numbers of bands for more sufficient training in data-insufficiency scenarios. 
To achieve it, (a) we first introduce a parallel Spectral-Spatial Separable Convolution (S3Conv) unit that efficiently extracts more spatial-spectral meaningful features than previous 3D \cite{wei20203} and separable convolutions \cite{dong2019deep}.  
(b) With the stronger local spectral-spatial inductive bias, the extracted features are then processed by a newly proposed Guided Spectral Self-Attention (GSSA) that performs the global self-attention along the 3D spectral rather than spatial \cite{pang2022trq3dnet} or 2D spectral/channel dimensions \cite{li2022spatial,cai2022mask} to selectively aggregate the information across different bands. This not only enriches our model with powerful capabilities for identifying long-range spectral correlations but also makes our model free of inflexibility of 2D spectral SA for dealing with different HSIs, quadratic complexity of spatial SA \cite{dosovitskiy2020image}, the issue of vanished long-range dependency of QRNN \cite{wei20203}. 
(c) Besides, inspired the relatively stationary global patterns and statistics of features of different HSI bands, as shown in \figref{fig:stat}, we propose to enhance our GSSA with a set of learnable queries that encode the global spectral statistics for each band. We alternatively switch between self-attention among spectral bands and cross-attention between spectral bands and learnable queries during the training, so that we can guide the GSSA to pay attention to features that are more discriminative and beneficial for denoising while keeping the flexibility. 
(d) Moreover, we propose a Self-Modulated Feed-Forward Network (SM-FFN) with a novel SM-branch to further strengthen the aggregated features of more informative regions. 
Extensive experiments on various datasets under different noise show that our HSDT consistently outperforms the existing state-of-the-art (SOTA) methods while maintaining low computational overhead, as shown in \figref{fig:flops}. 

In summary, our contributions are that, 
\begin{itemize}[noitemsep,topsep=0pt,leftmargin=*]
    \item We present HSDT, a 3D hybrid spectral denoising transformer that effectively captures the local spatial-spectral features and long-range global spectral correlations. 
    \item We introduce GSSA guided by a set of learnable queries that encode the global statistics of HSIs, which models long-range spectral correlations along 3D spectrum instead of previous 2D spectral/channel dimensions. 
    \item We propose SM-FFN with a novel self-modulated branch for driving the model to pay attention to more informative regions, along with a S3Conv for extracting spatial-spectral meaningful features.
\end{itemize}

%% file: section/related_work.tex
\section{Related Works}

\subsection{Hyperspectral Image Denoising}

Traditional approaches for HSI denoising usually formulate the task as an optimization problem, which is solved by imposing different types of handcrafted regularizations \cite{yuan2012hyperspectral,othman2006noise,sun2017hyperspectral,maggioni2012nonlocal, deltaprox2023}.  
Among these optimization-based methods, non-local similarity \cite{maggioni2012nonlocal} has been widely utilized for its ability to integrate the image patches across the spectral and spatial locations. 
To reduce the computational burden, global spectral low-rank correlation \cite{wei2019low,sun2017hyperspectral,zhao2020fast} has also been heavily studied. 
Besides, different enhanced total variation priors \cite{pengE3DTV,wang2017hyperspectral,yuan2012hyperspectral} are also adopted by considering the smoothness of local image patches. 
Though these methods could achieve favorable performance, most of them are computationally inefficient and can only address the noise satisfying the required assumptions, \eg, Gaussian noise. 

Meanwhile, recent works \cite{yuan2018hyperspectral,wei20203,lai2022deep, lai2023mixed} tend to exploit deep learning to learn denoising mapping purely in a data-driven manner. 
For most of these methods, the encoder-decoder U-Net \cite{ronneberger2015u, dong2019deep,lai2022deep,wei20203} architecture is the prominent choice due to its effectiveness for retaining both high- and low-level multi-scale representations. 
Residual learning \cite{he2016deep} is also widely adopted to reduce learning difficulties from different perspectives, \eg, residual image \cite{chang2018hsi} and residual features \cite{yuan2018hyperspectral,lai2022deep}. 
To consider the properties of HSIs, \eg, spatial-spectral correlations, QRNN3D \cite{wei20203} proposes to use 3D convolution and quasi-recurrent unit \cite{bradbury2016quasi}.  Our work adopts techniques, residual learning, 3D convolution, and U-shape architecture, but our blocks, \eg, S3Conv is more efficient than 3D convolution, and our GSSA could prevent vanished correlations for long-range spectral bands of QRU \cite{wei2019low}.
 Separable convolution \cite{howard2017mobilenets} is first introduced to replace 2D convolution. For HSI denoising, it is also adopted in \cite{imamura2019zero, dong2019deep, he2022spectrum} to reduce computational burden with similar motivations as ours. As a new alternative, our S3Conv is separable 3D instead of 2D convolution \cite{imamura2019zero} and more effective than previous 3D variant \cite{dong2019deep}.

\subsection{Vision Transformers} 

Transformer \cite{vaswani2017attention} has been first introduced as a parallel and purely attention-based alternative for recurrent neural networks \cite{chung2014empirical,hochreiter1997long} in the literature of natural language processing. Though it is originally designed for modeling text, recent works such as ViT \cite{dosovitskiy2020image} and DeiT \cite{touvron2021training}, have successfully transferred the transformer for high-level vision tasks.
 Recognizing the powerful representation abilities, this architecture is also expeditiously adapted for low-level tasks \cite{liang2021swinir, chen2021pre,li2023efficient}, such as natural image denoising. 
 Among these methods, one of the key problems they attempt to overcome is the quadratic complexity of the Self-Attention (SA) mechanism in the transformer. 
 To address it, SwinIR \cite{liang2021swinir} is proposed as an adaption of Swin transformer \cite{liu2021swin} that replaces global attention with a more efficient shift-window-based attention. Similarly, Uformer \cite{wang2022uformer} performs attention over non-overlapped patches and adopts U-Net architecture \cite{ronneberger2015u} to further increase efficiency. From a different perspective, Restormer \cite{zamir2022restormer} explores self-attention along the feature channels to realize the linear complexity. 
Despite their superior performance for various natural image restoration tasks, direct transfer of them for HSI can result in performance degradations since none of them consider the properties of HSI. 
Instead, our HSDT introduce S3Conv and GSSA that can extract more spectral correlated features, which is more suitable for HSI.  

For HSI processing, the applications of transformer previously concentrate more on the classification \cite{hong2021spectralformer,he2021spatial,qing2021improved} and spectral reconstruction \cite{cai2022mask,lin2022coarse}, but also recently extend to the HSI denoising \cite{pang2022trq3dnet, li2022spatial, li2023spectral, yu2023dstrans, chen2022hider}. For example, to exploit the spatial attention, TRQ3DNet \cite{pang2022trq3dnet} combines the QRU \cite{wei20203} and Uformer \cite{wang2022uformer} block, while SST \cite{li2022spatial} and  DSTrans \cite{yu2023dstrans} employ the Swin transformer block. To consider spectral correlations, most existing HSI transformers, including SST \cite{li2022spatial}, DSTrans \cite{yu2023dstrans}, and MST \cite{cai2022mask}, utilize a 2D spectral/channel attention similar to Restormer \cite{zamir2022restormer}. Basically, these methods have three major drawbacks including (i) the spatial attention, \eg, Uformer \cite{wang2022uformer} and Swin \cite{liu2021swin}, neither can model spectral relationships nor be computationally friendly for HSI, (ii) spectral attention as Restormer \cite{zamir2022restormer} and MST \cite{cai2022mask} essentially perform attention on feature channel dimension $C$ of 2D data $\mathbf{X} \in \R^{H\times W\times C}$ and $C$ could change for different layers, which makes them no longer spectral attention. (iii) Built based on 2D architectures for color images, these methods are usually not flexible for handle different HSIs in a single model. On the contrary, our GSSA is both effective for capturing global spectral correlations and flexible for diverse HSIs as we consider exact spectral attention along spectral $D$ dimension of 3D data $\mathbf{X} \in \R^{H\times W\times D \times C}$. 
Concurrent to our work, Hider \cite{chen2022hider} also consider 3D spectral attention. Different from it, our GSSA employs a simple pooling strategy instead of conv-reshape strategy as Restormer \cite{zamir2022restormer} to compute the query and key as well as learnable query, which makes it much more effective and efficient.

%% file: section/method.tex
\begin{figure*}[h]
\centering
\includegraphics[width=1\linewidth]{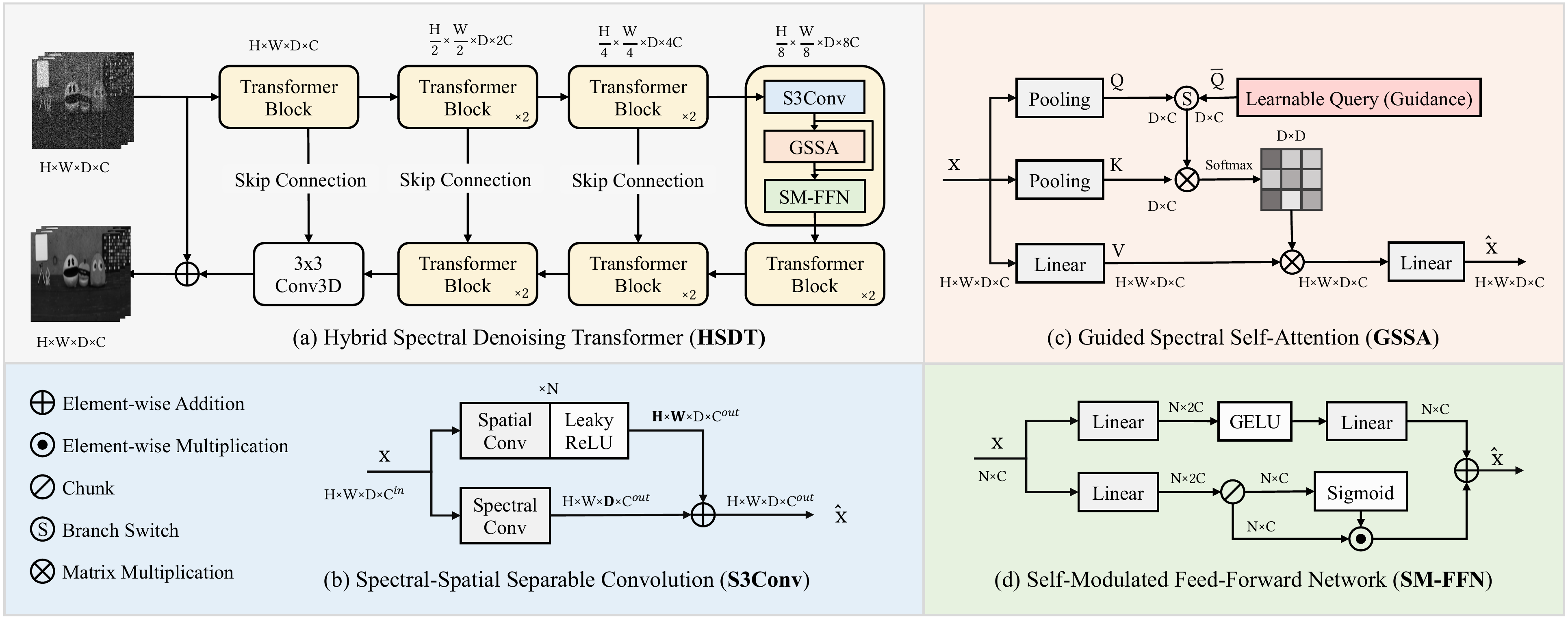}
\vspace{-5mm}
\caption{\textbf{Our HSDT architecture.} We adopt a hierarchical multiscale encoder-decoder (a) with each building transformer block stacks (d) a Spectral-Spatial Separable Convolution, (b) a Guided Spectral Self-Attention, and (c) a Self-Modulated Feed-Forward Network sequentially. We append batch normalization after each convolution and predict the residual image. 
}
\label{fig:arch}
\vspace{-2mm}
\end{figure*}

\section{Hybrid Spectral Denoising Transformer}

In this section, we present Hybrid Spectral Denoising Transformer (HSDT), a unified model for hyperspectral image denoising with an arbitrary number of bands. To achieve it effectively, our HSDT introduces several key designs, including (i) a powerful and lightweight spectral-spatial separable convolution as an alternative to 3D convolution, (ii) a guided spectral self-attention piloted by a set of learnable queries, and (iii) a self-modulated feed-forward network with an adaptive self-modulated branch.

The overall architecture of HSDT follows a U-shaped encoder-decoder with skip-connections \cite{ronneberger2015u}, which is depicted in \figref{fig:arch}\textcolor{linkcolor}{(a)}. Such hierarchical multi-scale design not only reduces the computational burden but also increases the receptive fields, which is different from conventional plain transformers \cite{liang2021swinir,dosovitskiy2020image}. In general, HSDT is built by stacking a series of transformer blocks as, 
\begin{align}
    \mathbf{\hat{X}} &= \operatorname{BN}(\operatorname{S3Conv}(\mathbf{X})) \\
    \mathbf{Y} &= \operatorname{SM-FFN}(\operatorname{GSSA}(\mathbf{\hat{X}}) + \mathbf{\hat{X}}) 
\end{align}
where $\mathbf{X} \in \R^{H\times W\times D \times C}, \mathbf{Y} \in \R^{\hat{H}\times \hat{W} \times D \times \hat{C}}$ are the input and output feature maps, $H,W$ denote spatial size, $D$ denotes the number of spectral bands, $C$ denotes the number of feature maps, and $\operatorname{BN}$ denotes batch normalization \cite{ioffe2015batch}. More specifically, given the input noisy HSI, it is first projected into low-level features through a head transformer block and then passed through several transformer blocks to fuse the features along both spatial and spectral dimensions. The residual connection is added to the final output and the input noisy image.  We use trilinear interpolations for upsampling and adopt additive skip connections in all levels of transformer blocks. Next, we illustrate the details of each network component.

\subsection{Spectral-Spatial Separable Convolution}

Modern HSI cameras are capable of capturing images with an exceedingly larger number of spectral bands, \eg, 31 for Specim PS Kappa DX4~\cite{fu2021biqrnn3d}, and 224 for AVIRIS sensor~\cite{kalman1997classification}. However, it is still difficult to collect a large amount of training data for each device due to the complex and time-consuming imaging processes. Hence, it is one of the major concerns for an HSI denoising method if it can handle images with a different number of bands using a single model so that the dataset collected by different devices can be jointly used for more sufficient training. 

To address the issue, the sliding-window denoising strategy \cite{yuan2018hyperspectral} and Conv3D \cite{wei20203} have been adopted to build band-flexible networks. However, their performance and efficiency are still limited by the local receptive field and heavy computation. In this work, we seek to preserve band flexibility and augment our denoising transformer with inductive bias while avoiding splitting HSI into windows. For this, we propose {Spectral-Spatial Separable Convolution} (S3Conv), a more lightweight and powerful variant of Conv3D 
that parallel applies spectral and spatial convolution.   

In detail, standard Conv3D filters spectral-spatial correlated features but introduces a heavy burden on the number of parameters and computation.
To alleviate the computational burden, our S3Conv decouples the Conv3D into two parallel branches, which separately process the inputs along the spatial and spectral dimensions, as illustrated in  \figref{fig:arch}\textcolor{linkcolor}{(b)}. The spatial convolution extracts features with 2D filters for each band and the spectral convolution applies $1\times 1$ projection to correlate the spectral information of all bands. To obtain the final features, we combine the output from two branches through element-wise addition. 
Instead of a direct extension of 2D separable Conv for 3D data, our S3Conv is spatial-spectral instead of spatial-channel separable, which makes it more suitable for extracting spectral correlation. Besides, our S3Conv is parallel instead of sequentially separated, which makes it easily accelerated.

\subsection{Guided Spectral Self-Attention}

Despite the spatial self-attention \cite{liang2021swinir,ronneberger2015u} improves the model performance by considering spatial interaction and non-local similarities, it is computationally demanding and might be difficult to deal with HSIs with a different number of bands. In this work, we propose an efficient Guided Spectral Self-Attention (GSSA) that applies 3D SA along the spectral than spatial nor channel dimensions. Our GSSA is intuitively supported by the spectral correlations of HSI and has linear complexity and long-range relation modeling abilities. This makes our model extremely more powerful at locating the informative regions to assist the denoising than existing spectral integration techniques \cite{wei20203, lai2022deep}.

\vspace{-2mm}
\paragraph{Spectral Self Attention.} GSSA takes 3D feature maps from previous S3Conv, \ie, $\mathbf{X} \in \R^{H \times W \times D \times C}$, and performs 3D attention on $D$ dimension, instead of $C$ dimension of $\mathbf{X} \in \R^{H \times W \times C}$ of previous 2D spectral attention \cite{cai2022mask,li2022spatial}. To perform attention, we first convert it into query, key, and value. Unlike conventional attention block \cite{vaswani2017attention}, only value  $\mathbf{\hat{V}} \in \R^{H \times W \times D \times C}$ is linearly projected from $\mathbf{X}$ in GSSA. Linear projections for query and key are not necessary according to our experiments, so we omit it for simplicity. Instead, we directly perform the global average pooling on input $\mathbf{X}$ along the spatial dimensions to obtain the global features of each band, \ie, $\mathbf{Q},\mathbf{K} \in \R^{D \times C}$. This pooling strategy is not only parameter-free but it also differs from previous reshape strategy \cite{zamir2022restormer,cai2022mask} that causes larger computation in the following dot-product attention.

Then, the transposed attention map $\mathbf{A}$ in the shape of $\R^{D\times D}$ is obtained via dot-product between key $\mathbf{K}$ and query $\mathbf{Q}$ with softmax normalization. Finally, we multiply the attention map with the value $\mathbf{\hat{V}}$ to dynamically select the essential features across the spectrum for each band, \ie,
\begin{align}
&\operatorname{Attention}(\mathbf{Q},\mathbf{K}, \mathbf{\hat{V}})=\mathbf{\hat{V}} \cdot \operatorname{Softmax}(\mathbf{K} \cdot \mathbf{Q}) \label{eq:attn} \\
&\hat{\mathbf{X}}=\mathbf{W} \operatorname{Attention}(\mathbf{Q}, \mathbf{K}, \mathbf{\hat{V}})+\mathbf{X}. \label{eq:attn2}
\end{align}
To better transform the features, we perform another linear projection $\mathbf{W} \in R^{C\times C}$ on the fused features. Beside, we learn the residual features to stabilize the training. 

\begin{figure}
   \centering
   \includegraphics[width=1\linewidth]{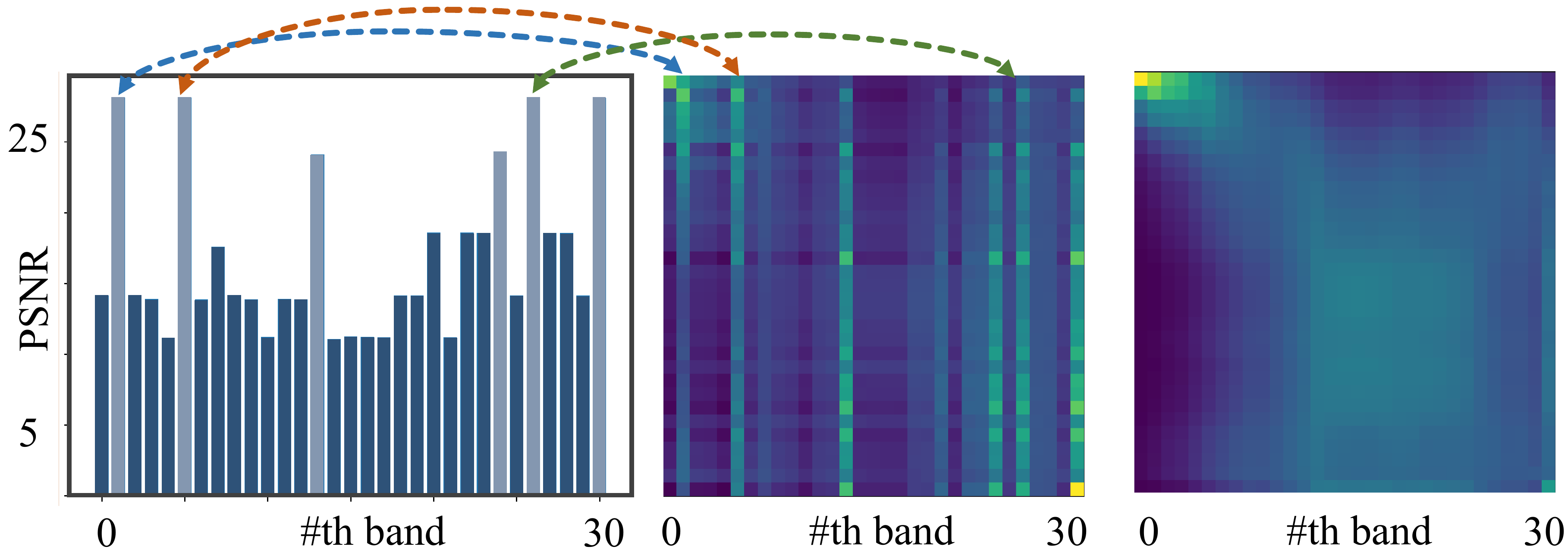}
    \begin{subfigure}[b]{0.135\textheight}
         \caption{SNR of each band}
     \end{subfigure}
     \hfill
     \begin{subfigure}[b]{0.13\textwidth}
         \caption{Attn w/ LQ}
     \end{subfigure}
     \hfill
     \begin{subfigure}[b]{0.13\textwidth}
         \caption{Attn w/o LQ}
     \end{subfigure}
   \vspace{-4mm}
   \caption{The Learnable Query (LQ) guides the model to pay attention (Attn) to more informative bands with higher signal-to-noise ratio (SNR). 
   }
   \label{fig:learnable_query}
   \vspace{-2mm}
\end{figure}

\vspace{-2mm}
\paragraph{Learnable Query.} Different spectral bands of HSIs often exhibit some fixed relative relationships due to the physical spectral constraints. Inspired by this, we investigate the global spectral feature maps of different HSIs, as shown in \figref{fig:stat}. Perhaps surprisingly, we found that clean and noisy HSIs exhibit clearly different shared patterns, which indicates the possibilities to utilize them for identifying more useful bands in GSSA. 
To achieve it, we propose to model these global statistics of each band with a set of learnable queries, as shown in \figref{fig:arch}\textcolor{linkcolor}{(c)}.
We perform cross-attention (CA) between the pooled keys and learnable queries to filter out most discriminative features for each band. 
The learnable queries are jointly trained with the other part of the model and not restricted to the input feature maps so that they can better fit the statistics of clean HSIs. As a results, the attention with learnable queries produce a remarkably better attention map with clear interpretability that identifies more informative bands with higher SNR to assist the denoising of more noisy ones, as shown in \figref{fig:learnable_query}. 

\vspace{-2mm}
\paragraph{Alternative Training Strategy.} 
The learnable queries are very useful to generate more discriminative attention, but the number of queries has to be predefined to the number of bands of the training dataset. It would disable the model from handling HSIs with a different number of bands simultaneously. To address the issue, we propose an alternative training strategy, in which we randomly switch between SA and CA with learnable queries during the training. Since the pooling operation in SA also captures global information to some extent, the proposed learnable query with this strategy can then be viewed as guidance leading the training of self-attention on more descriptive bands, which consequently leads to better performance.

\subsection{Self-Modulated Feed-Forward Network} 

\begin{figure}
     \centering
     \begin{subfigure}[b]{0.108\textheight}
         \centering
         \includegraphics[width=\textwidth]{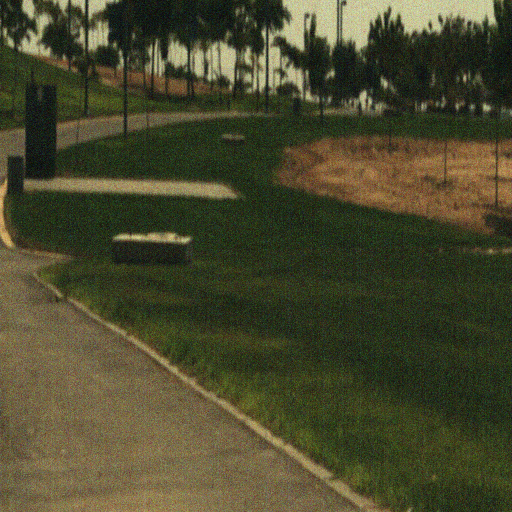}
         \caption{Noisy HSI}
     \end{subfigure}
     \hfill
     \begin{subfigure}[b]{0.108\textheight}
         \centering
         \includegraphics[width=\textwidth]{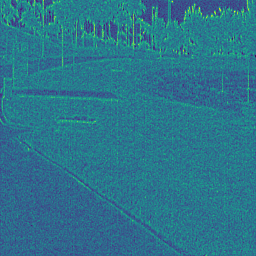}
         \caption{Feature map}
     \end{subfigure}
     \hfill
     \begin{subfigure}[b]{0.13\textheight}
         \centering
         \includegraphics[width=\textwidth]{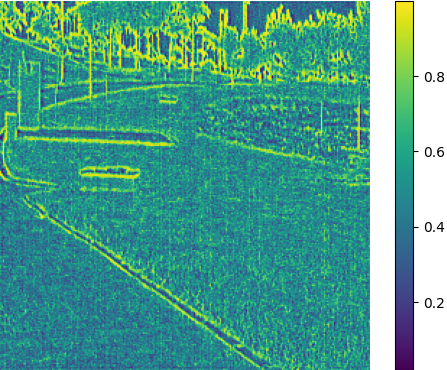}
         \caption{Modulation Weight}
     \end{subfigure}
      \vspace{-2mm}
   \caption{Our Self-Modulated FFN amplifies the features in high-information-density regions, \eg, edges, with an element-wise modulation weight. 
   Yellow regions denote higher weight.
   }
   \label{fig:smffn}
   \vspace{-2mm}
\end{figure}

Feed-Forward Network (FFN) is one of the most essential parts of transformer architectures, and it has been reported that it might be the key to construct the meta structure of transformer than SA \cite{yu2022metaformer}. Traditional FFN \cite{vaswani2017attention} processes the output features from the SA layer with two linear projections and a non-linear activation between them. 

\input{figures/gaussian_70_0308}
\input{tables/gaussian.tex}

In this work, we propose Self-Modulated FFN (SM-FFN) with a fundamental augmentation of the vanilla FFN using self-modulation. As shown in \figref{fig:arch}\textcolor{linkcolor}{(d)}, we first expand the input features channels $\mathbf{X} \in \R^{H\times W\times D \times C}$ with a scale factor of 2 through a linear projection, $\mathbf{Y}=\mathbf{W_3}\mathbf{X}, \mathbf{W_3} \in \R^{C\times 2C}$, then we split (also known as \emph{chunk} operation) the expanded features $\mathbf{Y}$ into two parts $\mathbf{F}, \mathbf{W} \in \R^{H\times W\times D \times C}$. We treat one part $\mathbf{F}$ as the candidate features and another part $\mathbf{W}$ after the sigmoid normalization as an element-wise modulation weight.
 With the vanilla FFN branch, our SM-FFN can be described as,
\begin{equation}
\operatorname{SM-Branch}(\mathbf{X}) = \mathbf{F} \odot \operatorname{Sigmoid}(\mathbf{W})  \label{eq:ffn},
\end{equation}
\begin{equation}
\operatorname{SM-FFN}(\mathbf{X}) = \mathbf{W_1}(\operatorname{GELU}(\mathbf{W_2} \mathbf{X})) + \operatorname{SM-Branch}(\mathbf{X}), \label{eq-sm2ffn}    
\end{equation}
where $\mathbf{W_2} \in \R^{C\times 2C}, \mathbf{W_1} \in \R^{2C\times C}$.
Intuitively, our SM branch is designed and works as a soft max-pooling that amplifies the activation of regions with higher information density. As reflected in \figref{fig:smffn}, this makes our network more robust by emphasizing the features in regions that are more important for improving the reconstruction quality, \eg, edges and corners.

%% file: figures/gaussian_70_0308.tex
\begin{figure*}
	\centering
	\small
	\hspace{-23mm}
	\setlength{\tabcolsep}{0.1cm}
	\begin{minipage}[l]{0.22\linewidth}
		\begin{flushleft}
			\vspace{-0.1mm}
			\hspace{-3mm}
			\includegraphics[width=0.96\linewidth]{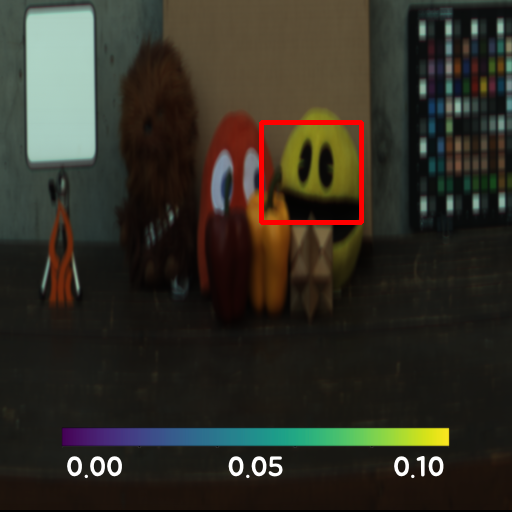}\\
			\begin{tabular}[c]{@{}c@{}}\hspace{10mm}Ground Truth\end{tabular}
		\end{flushleft}
	\end{minipage}
	\hspace{-5mm}
	\begin{minipage}[t]{0.5\linewidth}
		\begin{tabular}{ccccccc}
			\includegraphics[width=0.185\linewidth]{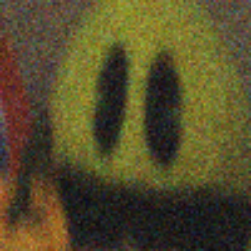}
			& \includegraphics[width=0.185\linewidth]{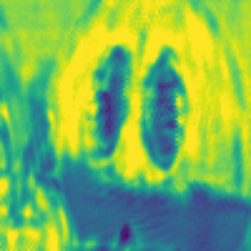}
			& \includegraphics[width=0.185\linewidth]{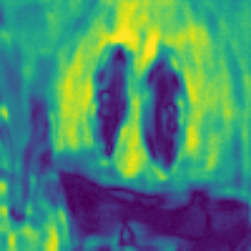}
			& \includegraphics[width=0.185\linewidth]{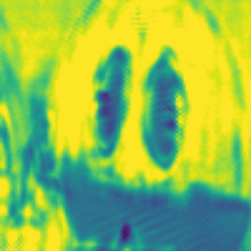}
			& \includegraphics[width=0.185\linewidth]{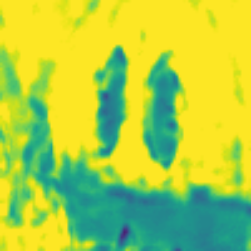}
			& \includegraphics[width=0.185\linewidth]{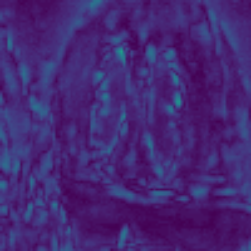}
			& \includegraphics[width=0.185\linewidth]{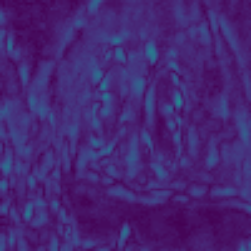}
			\\
			Noisy & LLRT \cite{chang2017hyper} & KBR\cite{Qi2017Kronecker}  & WLRTR\cite{chang2020weighted} & NGmeet\cite{he2019non}  & HSIDCNN\cite{yuan2018hyperspectral} & QRNN3D\cite{wei20203} \\
			
			\includegraphics[width=0.185\linewidth]{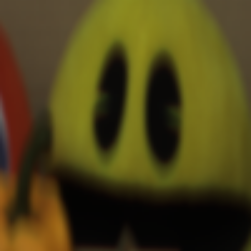}
			& \includegraphics[width=0.185\linewidth]{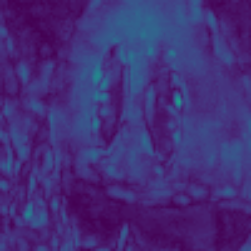}
			& \includegraphics[width=0.185\linewidth]{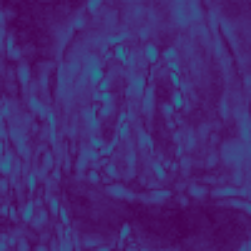}
			& \includegraphics[width=0.185\linewidth]{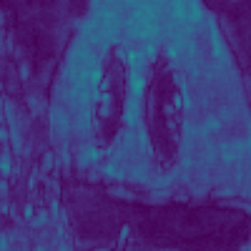}
			& \includegraphics[width=0.185\linewidth]{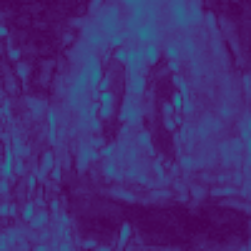}
			& \includegraphics[width=0.185\linewidth]{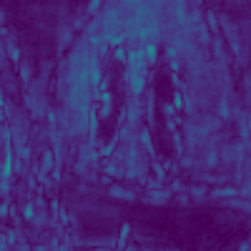}
			& \includegraphics[width=0.185\linewidth]{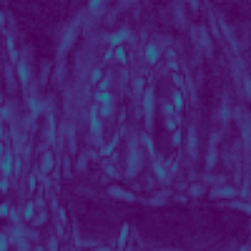}
			
			\\      
			 Ground Truth & T3SC\cite{bodrito2021trainable} & GRUNet\cite{lai2022deep} & Restormer\cite{zamir2022restormer} & TRQ3D\cite{pang2022trq3dnet} & SST \cite{li2022spatial} & HSDT (Ours)
		\end{tabular}
	\end{minipage}
	\hspace{22mm}

	\vspace{-3mm}
	\caption{Visual comparison with error maps for simulated Gaussian noise removal on ICVL dataset. ($\sigma=70$)}
	\label{fig:gaussian}
\end{figure*}

%% file: tables/gaussian.tex
\begin{table*}[ht]
   \small
   \setlength{\tabcolsep}{0.107cm}
   \begin{center}
   \begin{tabular}{|l|c|c|ccc|ccc|ccc|ccc|}
      \hline
      \rowcolor{graycolor}
        &              &          & \multicolumn{3}{c|}{$\sigma=30$} & \multicolumn{3}{c|}{$\sigma=50$} & \multicolumn{3}{c|}{$\sigma=70$} & \multicolumn{3}{c|}{$\sigma=\text{blind}$} \\ 
      \cline{4-15} 
      \rowcolor{graycolor}
      \multirow{1}{*}{Methods}           & Params (M)   &Runtime (s)      & PSNR     & SSIM     & SAM     & PSNR     & SSIM     & SAM     & PSNR     & SSIM     & SAM     & PSNR      & SSIM     & SAM      \\ \hline
Noisy             & -        & -        & 18.59    & 0.110    & 0.807   & 14.15    & 0.046    & 0.991   & 11.23    & 0.025    & 1.105   & 17.34     & 0.114    & 0.859    \\
LLRT \cite{chang2017hyper}             & -        & 627        & 41.99    & 0.967    & 0.056   & 38.99    & 0.945    & 0.075   & 37.36    & 0.930    & 0.087   & 40.97     & 0.956    & 0.064    \\
KBR \cite{Qi2017Kronecker}              & -        & 1755     & 41.48    & 0.984    & 0.088   & 39.16    & 0.974    & 0.100   & 36.71    & 0.961    & 0.113   & 40.68     & 0.979    & 0.080    \\ 
WLRTR \cite{chang2020weighted}            & -        & 1600     & 42.62    & 0.988    & 0.056   & 39.72    & 0.978    & 0.073   & 37.52    & 0.967    & 0.095   & 41.66     & 0.983    & 0.064    \\ 
NGmeet \cite{he2019non}           & -        & 166      & 42.99    & 0.989    & 0.050   & 40.26    & 0.980    & 0.059   & 38.66    & 0.974    & 0.067   & 42.23     & 0.985    & 0.053    \\ \hline
HSID-CNN \cite{yuan2018hyperspectral}         & 0.40     & 0.48     & 41.72    & 0.987    & 0.067   & 39.39    & 0.980    & 0.083   & 37.77    & 0.972    & 0.096   & 40.95     & 0.984    & 0.072    \\ 
QRNN3D  \cite{wei20203}           & 0.86     & 0.44     & 42.22    & 0.988    & 0.062   & 40.15    & 0.982    & 0.074   & 38.30    & 0.974    & 0.094   & 41.37     & 0.985    & 0.068    \\ 
T3SC  \cite{bodrito2021trainable}            & 0.83     & 0.95     & 42.36    & 0.986    & 0.079   & 40.47    & 0.980    & 0.087   & 39.05    & 0.974    & 0.096   & 41.52     & 0.983    & 0.085    \\ 
GRUNet \cite{lai2022deep}           & 14.2     & 0.87     & 42.84    & 0.989    & 0.052   & 40.75    & 0.983    & 0.062   & 39.02    & 0.977    & 0.080   & 42.03     & 0.987    & 0.057    \\ 
Restormer \cite{zamir2022restormer}        & 26.2     & 0.31     & 42.80    & 0.990    & 0.062   & 41.03    & 0.985    & 0.062   & 39.62    & 0.980    & 0.069   & 41.99     & 0.987    & 0.064    \\ 
TRQ3D \cite{pang2022trq3dnet}        & 0.68     & 0.47     & 43.25    & 0.990    & 0.046   & 41.30    & 0.985    & 0.053   & 39.86    & 0.980    & 0.061   & 42.47     & 0.988    & 0.054    \\ 
SST \cite{li2022spatial}        & 4.14     & 1.40     & 43.57    & 0.991    & 0.045   & 41.41    & 0.986    & 0.052   & 39.89    & 0.980    & 0.058   & 42.81     & 0.988    & 0.047    \\ 
SERT \cite{li2023spectral}        & 1.91    & 0.44     & 43.99    & 0.991    & 0.043   & 41.82    & 0.986    & 0.051   & 40.28    & 0.981    & 0.059   & 43.20     & 0.988    & 0.047    \\ 
\hline
HSDT-S(Ours)       & 0.13     & 0.36     & 43.31    & 0.990    & 0.047   & 41.16    & 0.985    & 0.055   & 39.66    & 0.980    & 0.064   & 42.57     & 0.988    & 0.051    \\
HSDT-M(Ours)       & 0.52     & 0.60     & \underline{44.02}    & \underline{0.991}    & \underline{0.041}   & \underline{41.82}    & \underline{0.986}    & \underline{0.049}   & \underline{40.33}    & \underline{0.981}    & \underline{0.055}   & \underline{43.32}     & \underline{0.989}    & \underline{0.045}    \\ 
HSDT-L(Ours)       & 2.09     & 0.69     & \textbf{44.31}    & \textbf{0.992}    & \textbf{0.041}   & \textbf{42.09}    & \textbf{0.987}    & \textbf{0.048}   & \textbf{40.59}    & \textbf{0.982}    & \textbf{0.054}   & \textbf{43.59}     & \textbf{0.989}    & \textbf{0.044}    \\ \hline

\end{tabular}
\end{center}
\vspace{-5mm}
   \caption{Gaussian denoising results on ICVL. \emph{Blind} denotes Gaussian noise with random sigma (ranged from 30 to 70). HSDT-M doubles the  width/channels of HSDT-S. HSDT-L increases the network depth of HSDT-M with an additional transformer block. The runtime is averaged over 10 runs.}
   \label{tab:denoise-gaussian}
   \vspace{-2mm}
\end{table*}

%% file: section/experiments.tex
\section{Experiments}

\subsection{Experimental Setup}

\input{figures/mixture_0308}

\input{tables/complex.tex}

\paragraph{Datasets.} We use three public HSI datasets with natural scenes, \ie, ICVL \cite{arad2016sparse}, CAVE \cite{CAVE_0293}, RealHSI \cite{zhang2021hyperspectral}, and one remotely sensed dataset Urban \cite{kalman1997classification}. ICVL and CAVE are unpaired datasets with solely clean images. RealHSI and Urban are paired and unpaired datasets with real-world noise. 
ICVL and CAVE share the same spectral resolution of 31, while RealHSI and Urban have different spectral resolutions of 34 and 210. We train and test the models on ICVL under simulated noise. For other datasets, we test the zero-shot performance with the models trained on ICVL. We use 100 images from ICVL for training, and we randomly select the main $512\times 512$ regions of 50 images from ICVL, and the entire regions of 12, 15 and 1 images from CAVE, RealHSI, and Urban for testing. 

\vspace{-4mm}
\paragraph{Noise Settings.} We consider simulated Gaussian and complex noise as well as real-world noise provided by RealHSI \cite{zhang2021hyperspectral} and Urban \cite{kalman1997classification}. The Gaussian noise is sampled from zero-mean i.i.d Gaussian distribution with different variances and we evaluate different methods on 30, 50, 70, and blind (range from 10 to 70) noise strengths. Following \cite{wei20203}, the complex noise is composed of non-i.i.d Gaussian noise and one or several types of complex noise, including, stripe, deadline, and impulse noise. 

\vspace{-4mm}
\paragraph{Implementation Details.} 
We implement the proposed method with PyTorch \cite{paszke2019pytorch}. 
The network is trained by optimizing the mean-square-root error between predicted images and the ground truth. We adopt Adam \cite{kingma2014adam} optimizer with a multi-step learning scheduler whose initial learning rate is set to $1\times 10^{-3}$ and decayed by a factor when it reaches the predefined milestones. Following \cite{wei20203}, we use a multi-stage training strategy to train models for Gaussian and complex noise. 
The learning rate warmup is used between training stages. 
The batch size and training patch size are set to 16 and $64\times 64$.
 All our models for Gaussian noise are trained for 80 epochs, and the models for complex noise are obtained by another 30 epochs of fine-tuning. 
We use pretrained complex denoising models for real noise.

\subsection{Main Results}
\noindent{\textbf{Gaussian Denoising.}}
We evaluate our method against SOTA optimization-based methods (\ie LLRT \cite{chang2017hyper}, KBR \cite{Qi2017Kronecker}, WLRTR \cite{chang2020weighted}, and NGmeet \cite{he2019non}), and deep-learning-based HSI denoising methods (\ie, HSID-CNN \cite{yuan2018hyperspectral}, QRNN3D \cite{wei20203}, GRUNet \cite{lai2022deep}, T3SC \cite{bodrito2021trainable}, TRQ3D \cite{pang2022trq3dnet}), SST \cite{li2022spatial}, and SERT \cite{li2023spectral}. The SOTA RGB denoising transformer, \ie, Restormer \cite{zamir2022restormer}, is also included by adjusting the input and output channels and training with the same HSI dataset as ours. 
As the quantitative results provided in \tabref{tab:denoise-gaussian}. we can observe that our method achieves superior performance outperforming the other ones by over 1 dB improvement on PSNR. Besides, we achieve the lowest SAM metric, which indicates that our method is better at maintaining spectral consistency. The synthetic color image is given in \figref{fig:gaussian} for the visual comparison.

\input{figures/remote_real2.tex}

\vspace{-2mm}
\paragraph{Complex Denoising.} 
While Gaussian denoising might be useful for some scenarios, \eg, as a plug-and-play denoiser \cite{lai2022deep}, it is not common for real-world images. We, therefore, also compare our method with several recently developed methods on simulated complex noise. 
As most optimization-based methods only perform well on the noise settings they can solve, we compare our method with a different set of optimization-based methods including, LRTV \cite{he2015total}, NMoG \cite{chen2017denoising} and TDTV \cite{wang2017hyperspectral}, in addition with the same set of deep-learning-based methods as Gaussian denoising.
As shown in \tabref{tab:denoise-complex}, our method again achieves the best performance with up to 1.7 dB improvement on PSNR. It demonstrates the stronger modeling ability of our method. The visual comparison is given in \figref{fig:complex}. 
 Similar to the results of Gaussian noise, our reconstruct better images.

\vspace{-2mm}
\paragraph{Real World Denoising.} 
We also conduct experiments on the recently developed RealHSI dataset \cite{zhang2021hyperspectral} using the models trained on ICVL. We compare our methods with the model in \cite{zhang2021hyperspectral} and several leading competing methods of complex denoising. The methods that cannot generalize to HSIs with 34 bands, \ie, T3SC \cite{bodrito2021trainable} and Restormer \cite{zamir2022restormer}, SST \cite{li2022spatial}, SERT \cite{li2023spectral}, TRQ3D \cite{pang2022trq3dnet}, are not included. 
The quantitative results are given in  \tabref{tab:real}. It can be seen that our method achieves better performance with fewer parameters.
The visual results are provided in the bottom row of \figref{fig:real}. It can be observed that our method produces cleaner and sharper results while the others are blurry or could not completely remove the noise. 

\input{tables/additional3.tex}

\subsection{Generalization to Other Datasets}

We have demonstrated the stronger generalization abilities of our method on real-world denoising dataset \cite{zhang2021hyperspectral} with models trained purely on ICVL and simulated noise. With the same models, we provide results on more datasets with different scenes and the number of bands. 

\vspace{-2mm}
\paragraph{Natural HSI.} The CAVE \cite{CAVE_0293} is another widely used HSI denoising dataset that contains more indoor scenes, \eg, different materials and objects. We evaluate the performance under mixture complex noise with the models trained on ICVL. As shown in \tabref{tab:real}, our model still achieves the best performance against the others with a large margin, which proves the stronger generalization capabilities of our method to handle out-of-distribution images.

\vspace{-2mm}
\paragraph{Remotely Sensed HSI.} We demonstrate that our model trained on ICVL with 31 bands can be directly transferred to datasets with a totally different number of bands, \eg, Urban with 210 bands, without any fine-tuning. This is largely supported by the flexibility of the proposed GSSA and S3Conv units, while such flexibilities are not presented in models, \eg, Restormer \cite{zamir2022restormer} and TRQ3D \cite{pang2022trq3dnet}. We present the visual comparison in \figref{fig:real}. It can be observed that QRNN3D \cite{wei20203} could not completely remove the row noise and lose the details of the roof. HSID-CNN \cite{yuan2018hyperspectral} and TDTV \cite{he2015total} eliminate the most noise but produce more blurry results. Our method instead removes most noise while maintaining the sharper details.

\input{tables/ablation_breakdown2.tex}

\subsection{Ablation Studies}

We adopt Gaussian denoising ($\sigma=50$) on ICVL to conduct the ablation studies. The baseline model is derived by removing our S3Conv, GSSA, and SM-FNN from HSDT-M. We also conduct the per-component ablation studies, which are provided in the supplementary material.

\vspace{-4mm}
\paragraph{Break-down Ablation.} 
We provide the results of break-down ablations in \tabref{tab:ablation-step}, in which we gradually add the proposed blocks back to the baseline model. It can be seen that GSSA provides the most performance gain, which can be attributed to the importance of spectral correlations for HSI denoising. The LQ guidance and SM-FFN improve the models with negligible parameter growth, while S3Conv improves the performance with even fewer parameters.

%% file: figures/mixture_0308.tex
\begin{figure*}
	\centering
	\small
	\hspace{-12mm}
	\setlength{\tabcolsep}{0.1cm}
	\begin{minipage}[l]{0.22\linewidth}
		\begin{flushleft}
			\vspace{-0.1mm}
			\hspace{-3mm}
			\includegraphics[width=1.045\linewidth]{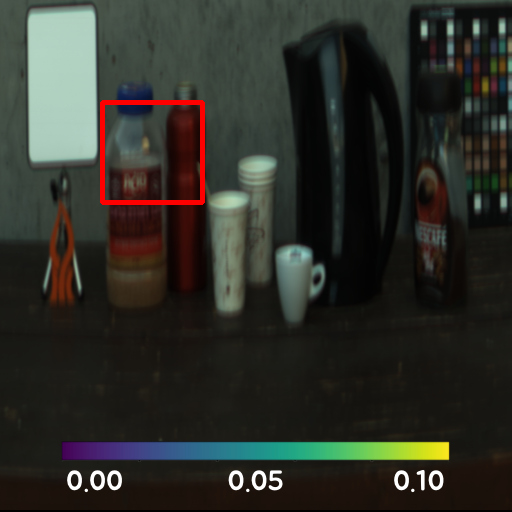}\\
			\begin{tabular}[c]{@{}c@{}}\hspace{7mm}Ground Truth\end{tabular}
		\end{flushleft}
	\end{minipage}
	\hspace{-1mm}
	\begin{minipage}[t]{0.5\linewidth}
		\begin{tabular}{cccccc}
			\includegraphics[width=0.2\linewidth]{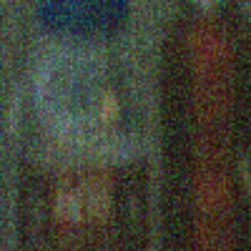}
			& \includegraphics[width=0.2\linewidth]{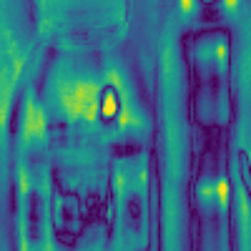}
			& \includegraphics[width=0.2\linewidth]{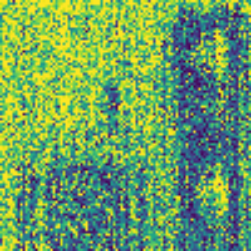}
			& \includegraphics[width=0.2\linewidth]{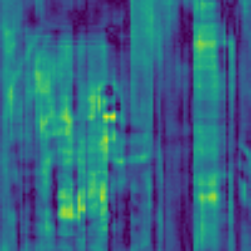}
			& \includegraphics[width=0.2\linewidth]{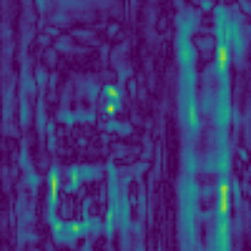}
			& \includegraphics[width=0.2\linewidth]{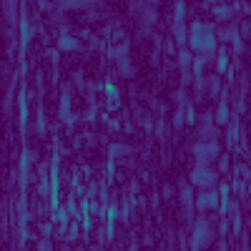}
			\\
			Noisy & LRTV \cite{he2015total} & NMoG \cite{chen2017denoising}  & TDTV \cite{wang2017hyperspectral} & HSID-CNN \cite{yuan2018hyperspectral}  & QRNN3D \cite{wei20203}\\
			
			\includegraphics[width=0.2\linewidth]{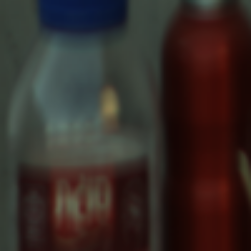}
			& \includegraphics[width=0.2\linewidth]{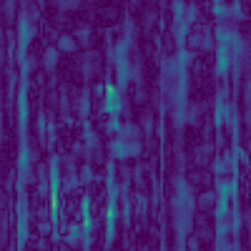}
			& \includegraphics[width=0.2\linewidth]{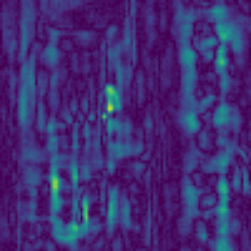}
			& \includegraphics[width=0.2\linewidth]{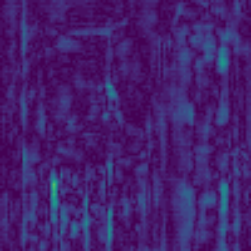}
			& \includegraphics[width=0.2\linewidth]{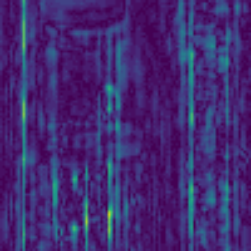}
			& \includegraphics[width=0.2\linewidth]{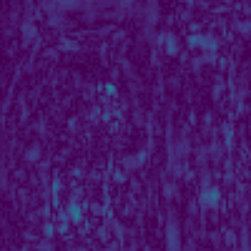}
			
			\\      
			 Ground Truth & GRUNet \cite{lai2022deep} & Restormer \cite{zamir2022restormer} & TRQ3D \cite{pang2022trq3dnet} & SST \cite{li2022spatial} & HSDT (Ours)
		\end{tabular}
	\end{minipage}
	\hspace{22mm}

	\vspace{-3mm}
	\caption{Visual comparison with error maps  for simulated complex noise removal on ICVL dataset. (\emph{mixture} case)}
	\label{fig:complex}
\end{figure*}

%% file: tables/complex.tex
\begin{table*}[t]
   \small
   \begin{center}
   \setlength{\tabcolsep}{0.126cm}
      \begin{tabular}{|l|ccc|ccc|ccc|ccc|ccc|}
         \hline
         \rowcolor{graycolor}
                & \multicolumn{3}{c|}{non-iid} & \multicolumn{3}{c|}{stripe} & \multicolumn{3}{c|}{deadline} & \multicolumn{3}{c|}{impulse} & \multicolumn{3}{c|}{mixture} \\ 
         \cline{2-16} 
         \rowcolor{graycolor}
         \multirow{1}{*}{Methods}            & PSNR     & SSIM     & SAM     & PSNR     & SSIM     & SAM     & PSNR     & SSIM     & SAM     & PSNR      & SSIM     & SAM   & PSNR      & SSIM     & SAM     \\ \hline
Noisy                & 18.25 & 0.168 & 0.898 & 17.80 & 0.159 & 0.910 & 17.61 & 0.155 & 0.917 & 14.80 & 0.114 & 0.926 & 14.08 & 0.099 & 0.944   \\
LRTV  \cite{he2015total}               & 33.62 & 0.905 & 0.077 & 33.49 & 0.905 & 0.078 & 32.37 & 0.895 & 0.115 & 31.56 & 0.871 & 0.242 & 30.47 & 0.858 & 0.287   \\ 
NMoG \cite{chen2017denoising}                & 34.51 & 0.812 & 0.187 & 33.87 & 0.799 & 0.265 & 32.87 & 0.797 & 0.276 & 28.60 & 0.652 & 0.486 & 27.31 & 0.632 & 0.513    \\ 
TDTV  \cite{wang2017hyperspectral}               & 38.14 & 0.944 & 0.075 & 37.67 & 0.940 & 0.081 & 36.15 & 0.930 & 0.099 & 36.67 & 0.935 & 0.094 & 34.77 & 0.919 & 0.113   \\ \hline
HSID-CNN \cite{yuan2018hyperspectral}            & 40.14 & 0.984 & 0.067 & 39.53 & 0.983 & 0.720 & 39.49 & 0.983 & 0.071 & 36.69 & 0.959 & 0.156 & 35.36 & 0.954 & 0.169   \\ 
QRNN3D \cite{wei20203}              & 42.79 & 0.978 & 0.052 & 42.35 & 0.976 & 0.055 & 42.23 & 0.976 & 0.056 & 39.23 & 0.945 & 0.109 & 38.25 & 0.938 & 0.108   \\ 
T3SC \cite{bodrito2021trainable}              & 41.28 & 0.987 & 0.065 & 40.85 & 0.986 & 0.072 & 39.54 & 0.983 & 0.096 & 36.06 & 0.952 & 0.203 & 34.48 & 0.946 & 0.228   \\ 
GRUNet \cite{lai2022deep}              & 42.89 & 0.992 & 0.047 & 42.39 & 0.991 & 0.050 & 42.11 & 0.991 & 0.050 & 40.70 & {0.985} & {0.067} & 38.51 & 0.980 & {0.081}   \\
Restormer \cite{zamir2022restormer}           & 40.81 & 0.987 & 0.050 & 40.49 & 0.986 & 0.052 & 39.89 & 0.986 & 0.055 & 37.60 & 0.972 & 0.746 & 36.21 & 0.968 & 0.752   \\ 
TRQ3D \cite{pang2022trq3dnet}           & 43.34 & 0.992 & 0.042 & 43.05 & 0.992 & 0.043 & 42.70 & 0.992 & 0.045 & 41.22 & \underline{0.983} & \underline{0.075} & 40.27 & {0.983} & \underline{0.075}   \\
SST \cite{li2022spatial}           & 43.43 & 0.993 & 0.042 & 43.02 & 0.992 & 0.044 & 42.95 & 0.992 & 0.044 & 41.27 & \textbf{0.985} & \textbf{0.064} & 39.19 & \textbf{0.983} & \textbf{0.067}   \\
SERT \cite{li2023spectral}           & 44.10 & 0.993 & 0.038 & 43.80 & 0.993 & 0.040 & 43.55 & 0.993 & 0.041 & 40.56 & 0.977 & 0.080 & 39.25 & 0.977 & 0.079   \\
\hline
HSDT-S(Ours)       & 43.46      & 0.992     & 0.044    & 43.13    & 0.992   & 0.047    & 42.97    & 0.991   & 0.047    & 41.11    & 0.976   & 0.110     &  40.22   & 0.974 & 0.116    \\
HSDT-M(Ours)       & \underline{44.56}     & \underline{0.994}     & \underline{0.039}    & \underline{44.29}    & \underline{0.993}   & \underline{0.040}  & \underline{44.18}     & \underline{0.993}    & \underline{0.040}  & \underline{41.28}    & 0.977   & 0.101    & \underline{40.46}    & 0.976 & 0.106  \\ 
HSDT-L(Ours)       & \textbf{44.94}     & \textbf{0.994}     & \textbf{0.036}    & \textbf{44.69}    & \textbf{0.994}   & \textbf{0.038}        & \textbf{44.55}   & \textbf{0.994}    & \textbf{0.037}    & \textbf{42.02}   & 0.980     & 0.101    & \textbf{41.07}   & \underline{0.980} & 0.101 \\ 
\hline
\end{tabular}
\end{center}
\vspace{-5mm}
\caption{Complex denoising results on ICVL. \emph{non-iid} denotes Non i.i.d Gaussian noise. \emph{stripe}, \emph{deadline}, \emph{impulse} denote the combination of \emph{non-iid} and corresponding complex noise. \emph{mixture} denotes the combination of all the mentioned noise. }
   \label{tab:denoise-complex}
   \vspace{-2mm}
\end{table*} 

%% file: figures/remote_real2.tex
\begin{figure*}
\captionsetup[subfigure]{labelformat=empty}
     \begin{center}
     \begin{subfigure}[b]{0.15\textwidth}
         \centering
         \includegraphics[width=\textwidth]{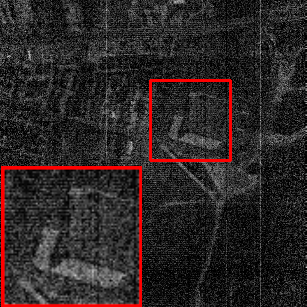}
         \caption{Noisy}
     \end{subfigure}
     \hfill
     \begin{subfigure}[b]{0.15\textwidth}
         \centering
         \includegraphics[width=\textwidth]{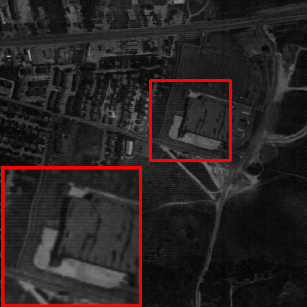}
         \caption{TDTV \cite{wang2017hyperspectral}}
     \end{subfigure}
      \hfill
     \begin{subfigure}[b]{0.15\textwidth}
         \centering
         \includegraphics[width=\textwidth]{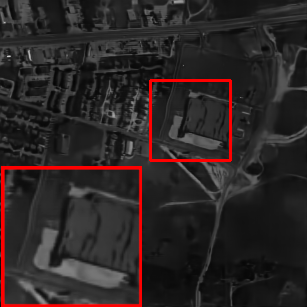}
         \caption{HSID-CNN \cite{yuan2018hyperspectral}}
     \end{subfigure}
      \hfill
     \begin{subfigure}[b]{0.15\textwidth}
         \centering
         \includegraphics[width=\textwidth]{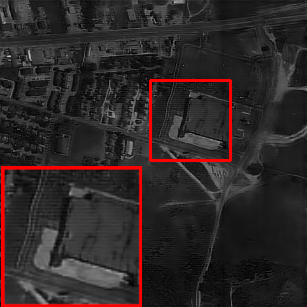}
         \caption{QRNN3D \cite{wei20203}}
     \end{subfigure}
      \hfill
     \begin{subfigure}[b]{0.15\textwidth}
         \centering
         \includegraphics[width=\textwidth]{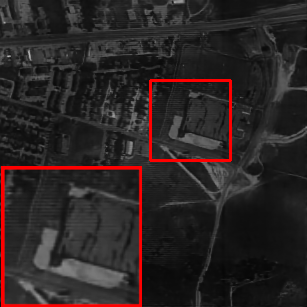}
         \caption{HSIDwRD \cite{zhang2021hyperspectral}}
     \end{subfigure}
      \hfill
     \begin{subfigure}[b]{0.15\textwidth}
         \centering
         \includegraphics[width=\textwidth]{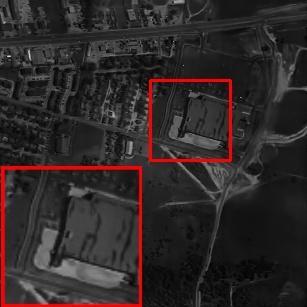}
         \caption{HSDT(Ours)}
     \end{subfigure}
     \quad
     \begin{subfigure}[b]{0.15\textwidth}
        \centering
        \includegraphics[width=\textwidth]{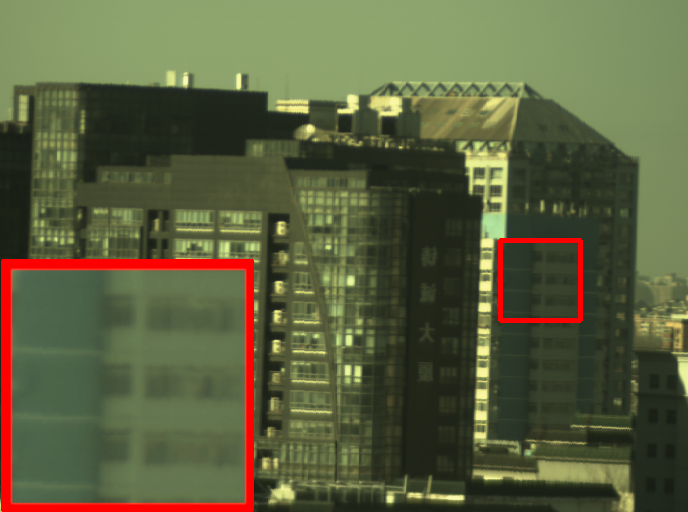}
        \caption{Ground Truth}
    \end{subfigure}
    \hfill
     \begin{subfigure}[b]{0.15\textwidth}
         \centering
         \includegraphics[width=\textwidth]{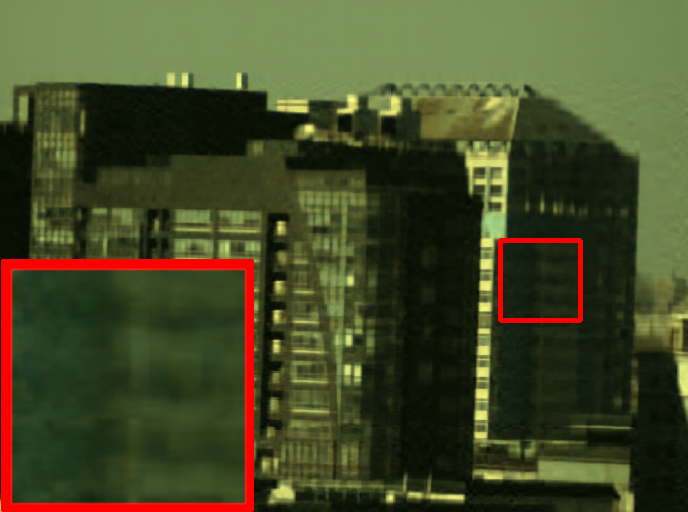}
         \caption{TDTV \cite{wang2017hyperspectral}}
     \end{subfigure}
     \hfill
     \begin{subfigure}[b]{0.15\textwidth}
         \centering
         \includegraphics[width=\textwidth]{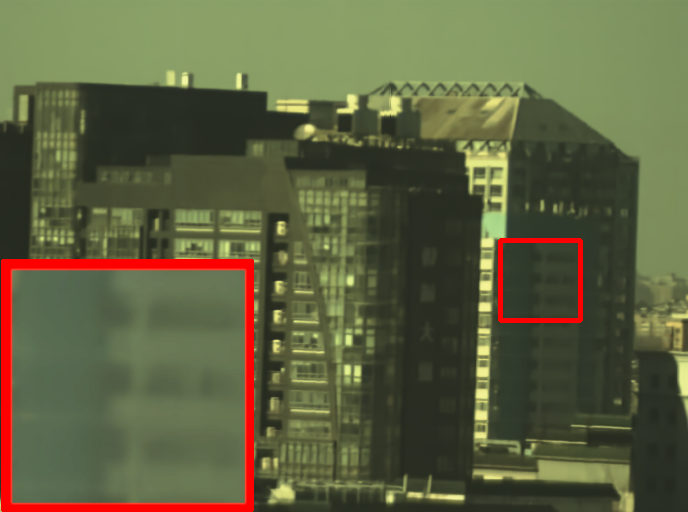}
         \caption{HSID-CNN \cite{yuan2018hyperspectral}}
     \end{subfigure}
     \hfill
     \begin{subfigure}[b]{0.15\textwidth}
         \centering
         \includegraphics[width=\textwidth]{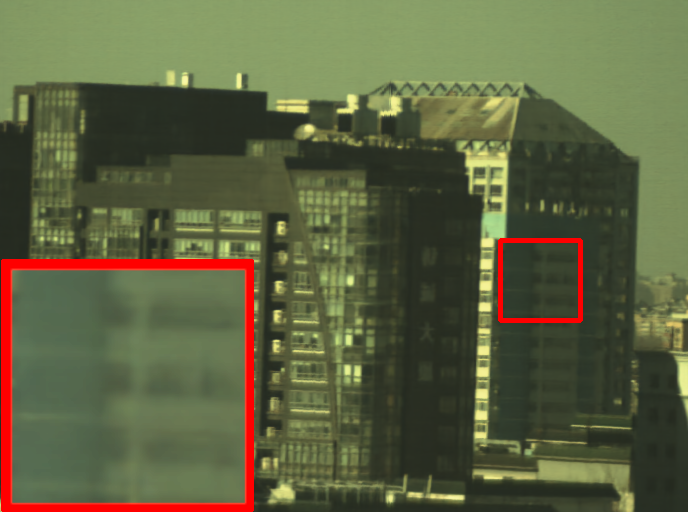}
         \caption{QRNN3D \cite{wei20203}}
     \end{subfigure}
      \hfill
     \begin{subfigure}[b]{0.15\textwidth}
         \centering
         \includegraphics[width=\textwidth]{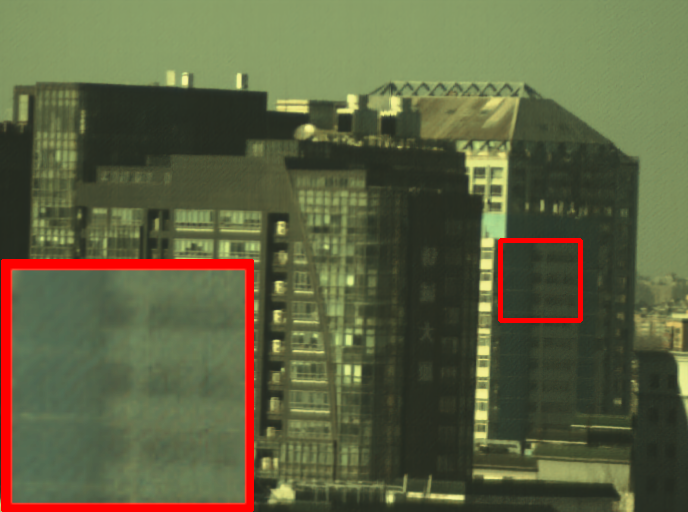}
         \caption{HSIDwRD \cite{zhang2021hyperspectral}}
     \end{subfigure}
     \hfill
     \begin{subfigure}[b]{0.15\textwidth}
         \centering
         \includegraphics[width=\textwidth]{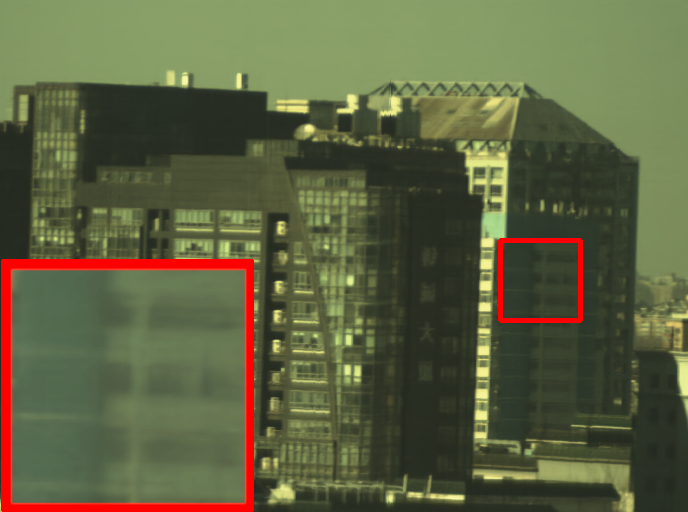}
         \caption{HSDT(ours)}
     \end{subfigure}
     \end{center}
     \vspace{-6mm}
   \caption{Visual results for real-world noise removal on Urban Dataset (Top) and Real \cite{zhang2021hyperspectral} Dataset (Bottom).}
   \label{fig:real}
   \vspace{-1mm}
\end{figure*}

%% file: tables/additional3.tex
\begin{table}[t]
\small
    \setlength{\tabcolsep}{0.132cm}
        \centering
        \small
        \begin{tabular}{|l |c | cc | cc|}
        \hline
        \rowcolor{graycolor}  &  & \multicolumn{2}{c|}{RealHSI \cite{zhang2021hyperspectral}} & \multicolumn{2}{c|}{CAVE \cite{CAVE_0293}} \\ 
        \cline{3-6} 
        \rowcolor{graycolor}
        \multirow{1}{*}{Methods}         & \multirow{1}{*}{Params(M)}    & PSNR          & SAM     & PSNR          & SAM     \\ \hline
	Noisy					&-	& 23.31 & 0.257 & 18.99 & 0.901\\
        NMoG \cite{chen2017denoising}  		 &-	& 30.90 & 1.762 & 30.84 & 37.86\\
        TDTV \cite{wang2017hyperspectral}  	 &-	& 31.14 & 1.853 & 33.14 & 22.34\\ 
        \hline
        HSID-CNN \cite{yuan2018hyperspectral}   &0.40 & 31.05 & 0.096 & 36.09 & 0.318\\
        QRNN3D \cite{wei20203} 			&0.86	& 31.13 & 0.094 & 37.80 & 0.247\\
        GRUNet \cite{lai2022deep} 	        &14.2	& 31.03 & 0.091 & 37.33 & 0.288\\
        HSIDwRD \cite{zhang2021hyperspectral}   &23.6 & 31.23 & 0.092 & 39.37 & 0.188\\
        \hline 
        HSDT-L(Ours)  			        &0.52 & \textbf{31.42} & \textbf{0.091} & \textbf{39.80} & \textbf{0.174}\\
        \hline
       \end{tabular}
       \vspace{-2mm}
       \caption{Additional results on RealHSI and CAVE. }
       \label{tab:real}
\end{table}

%% file: tables/ablation_breakdown2.tex
\begin{table}[t]
    \setlength{\tabcolsep}{0.12cm}
    \centering
    \small
    \begin{tabular}{|c  c  c  c | c | c | c|}
        \hline
        \rowcolor{graycolor} SSA        & Guidance   & SM-FFN     & Conv   & Params & PSNR           & SAM            \\
        \hline
        \xmark          &   \xmark         &   \xmark         & Conv3D & 0.43M  & 38.28          & 0.101          \\
        \cmark &   \xmark         &    \xmark        & Conv3D & 0.55M  & 41.34          & 0.057          \\
        \cmark & \cmark &   \xmark         & Conv3D & 0.55M  & 41.46          & 0.054          \\
        \cmark & \cmark & \cmark & Conv3D & 0.58M  & 41.62          & 0.052          \\
        \cmark & \cmark & \cmark & S3Conv & 0.52M  & \textbf{41.82} & \textbf{0.049} \\ \hline
    \end{tabular}
    \vspace{-2mm}
    \caption{Break-down ablation on the proposed components. }
    \label{tab:ablation-step}
    \vspace{-4mm}
\end{table}

%% file: section/conclusion.tex
\section{Conclusion}

We present HSDT, an effective and flexible transformer for hyperspectral image denoising. Built upon the hybrid hierarchical architecture, our HSDT is equipped with a novel S3Conv, GSSA, and SM-FFN module to effectively integrate the local spectral-spatial inductive bias and the long-range spectral dependency modeling.  Our S3Conv extracts highly correlated local spatial-spectral features without harming efficiency and flexibility, while the GSSA provides stronger capabilities for capturing global spectral correlations, guided by a set of learnable queries that encode the band-wise spectral signatures. With the SM-FFN to further strengthen the aggregated features of more informative regions, our model outperforms the existing SOTA methods on various datasets under simulated and real-world noise.

%% file: section/appendix.tex
\appendix

\section{More Details about GSSA}

\paragraph{Computational Complexity of GSSA.} Given an input $\mathbf{X} \in \R^{H \times W \times D \times C}$ where $H,W$ denote height and width,  $D$ denotes the number of spectral bands, $C$ denotes the features channels, the computational complexity of each step of GSSA is summarized in \tabref{tab:gssa-complexity}. Since the feature channels are typically larger than the number of spectral bands, the asymptotic computational complexity of GSSA is dominated by two linear transformations, \ie, \emph{Linear for $V$} and \emph{Post linear}.

\begin{table}[h]
\small
\centering
\begin{tabular}{l|l}
\hline
\textbf{Step} & \textbf{Complexity} \\ 
\hline
Linear for $V$     &   $(H \times W \times D) \times C^2$         \\
Pooling for $Q$,$K$     &  $ 2\times (H \times W) \times (D \times C)$          \\
Compute attention matrix     & $D\times D \times C$           \\ 
Feature aggregation    & $H \times W \times C \times D \times D$           \\ 
Post linear    & $(H \times W \times D) \times C^2$           \\ 
Total &  $O((H \times W \times D) \times C^2)$ \\
\hline
\end{tabular}
\caption{The computational complexity of GSSA. The overall complexity of GSSA is linear with respect to image size.}
\label{tab:gssa-complexity}
\end{table}

\vspace{-3mm}
\paragraph{Fast Implementation.} 

With the simplification of pixel-wise attention via global average pooling, our GSSA can be efficiently implemented with a depth-wise convolution by treating the shared attention map as a convolution filter and swapping the spectral and channel dimensions. The speed comparison is shown in \tabref{tab:gssa}, and it can be seen that the Conv-based implementation is approximately 20\% faster than the naive \verb+Matmul+-based one.

\begin{table}[h]
\setlength{\tabcolsep}{0.5cm}
\centering
\small
\begin{tabular}{l|c|c}
\hline
\textbf{Implementation}     &  \textbf{ Runtime (s)}  & \textbf{PSNR}    \\ \hline
Matmul-based       &  0.60 & 41.82 \\ 
Conv-based       &  0.47  & 41.82\\ \hline
\end{tabular}
\caption{Speed of different implementations of GSSA. Our Conv-based implementation reduces the running time without harming the performance.}
\label{tab:gssa}
\vspace{-3mm}
\end{table}

\begin{figure*}[t]
\centering
\includegraphics[width=1\linewidth]{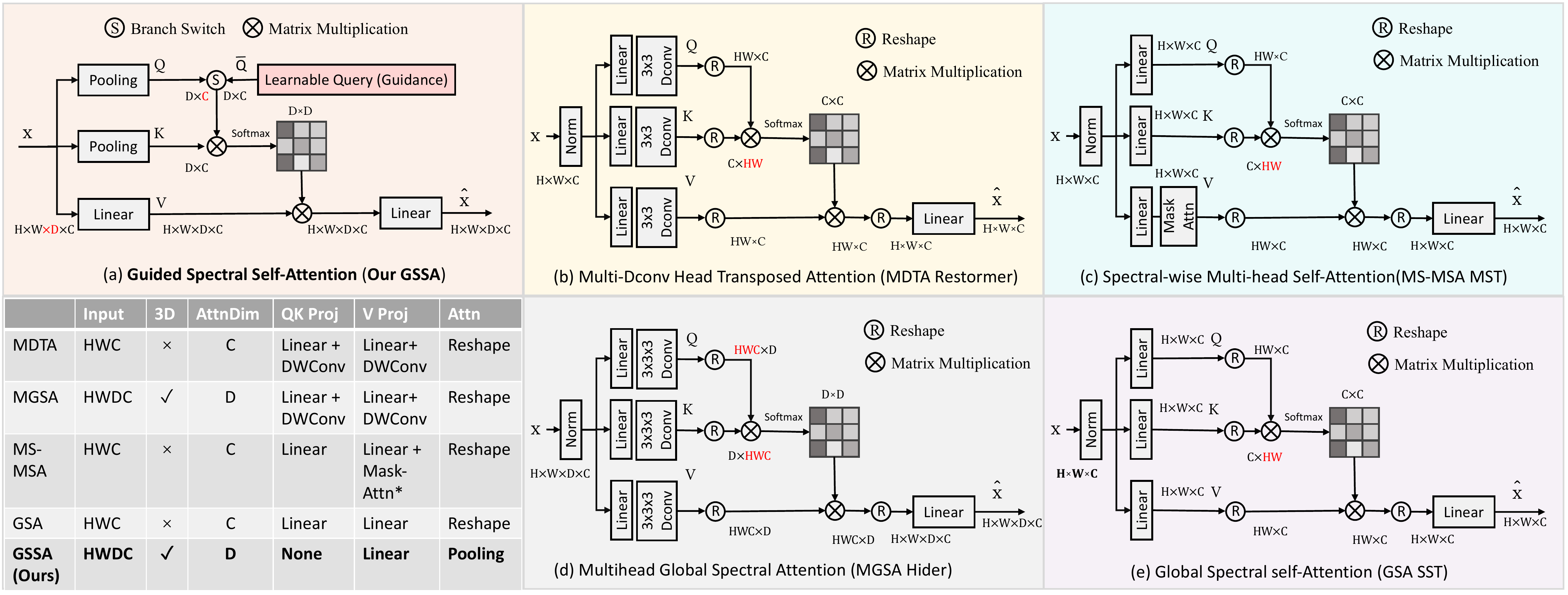}
\vspace{-5mm}
\caption{Comparison of different spectral/channel attention mechanisms. Our GSSA is significantly different from previous attention mechanisms. We could observe MDTA and MGSA are almost identical; MS-MSA and GSA are almost identical. Besides, MS-MSA and GSA are also basically simpler version of MDTA without depthwise convolution. Please refer to the text for detailed explanation. 
}
\label{fig:cmp}
\vspace{-2mm}
\end{figure*}

\vspace{-3mm}
\subsection{Comparison against other Attention.} 
Here, we provide a more detailed explanation regarding the differences between our GSSA and existing channel or spectral attention mechanisms.
\emph{We highlight that our GSSA is significantly different from previous attention mechanisms in a variety aspects}. Since GSSA performs attention along spectral rather than spatial dimensions, we here compare it with four previous attention mechanisms that apply along spectral or channel dimensions including: 
\begin{table}[h]
\small
\centering
\begin{tabular}{c|c|c}
\hline
\textbf{Attention} & \textbf{Method} &  \textbf{Task} \\ 
\hline
MDTA & Restormer \cite{zamir2022restormer} & Color image restoration \\
MS-MSA & MST \cite{cai2022mask} & Spectral Reconstruction \\
GSA & SST \cite{li2022spatial} & HSI denoising \\
MGSA & Hider \cite{chen2022hider} & HSI denoising \\
\hline
\end{tabular}
\caption{The competing attention mechanisms.}
\label{tab:comparsison}
\end{table}

\figref{fig:cmp} illustrates the structures of the aforementioned attention mechanisms. It is worth noting that all previous methods are essentially variants of MDTA proposed in Restormer, whereas our GSSA is fundamentally distinct from them. In the following, we will provide a detailed explanation of the main differences between the previous methods and our GSSA.

\textbf{3D vs 2D Data Format.} The first notable difference, which can be easily confused with previous work, is that \emph{our GSSA performs attention on the spectral dimension}, i.e., the $D$ dimension of a 5D data cube $x \in \R^{B\times C \times D \times H \times W}$. In contrast, previous works, such as MST, and SST, even though they refer to their attention mechanisms as spectral attention, essentially apply channel attention along the $C$ dimension of a 4D data cube $x \in \R^{B\times C \times H \times W}$, which is the same as MDTA.
Our 3D approach provides the flexibility to handle HSIs with different bands within a single model. Additionally, it achieves superior performance by preserving the structures of different bands, \ie, each band possesses its own feature set, and their relationship remains unchanged across layers of the entire model.

\textbf{QKV Projection.} The second key difference pertains to the projections used for the query, key, and value. Conventional attention mechanisms typically employ three linear projections to project the input into query, key, and value. This approach is utilized in all of the compared methods, with the exception of our GSSA. Instead, \emph{our GSSA applies linear projection solely for the value, which greatly simplifies the design}. By contrast, MDTA needs a extra 3x3 depthwise convolution after the linear projection. MGSA is identical to MDTA, except that it employs a 3D convolution. MS-MSA and GSA are the same and solely utilize linear projections, with the exception that MS-MSA employs an additional mask attention specifically designed for spectral reconstruction.

\textbf{Pooling vs Reshape.} The third difference is that our GSSA uses global average pooling to obtain feature maps for each band. This differs from previous methods that adopt a reshape approach. Our method is significantly more computationally efficient compared to previous approaches. Previous methods reshape the Q, K, and V tensors from a shape of $H \times W \times C$ into $HW \times C$, treating $HW$ as the features for each channel. This leads to a time complexity of dot-product attention that is linear with respect to the image size, \ie, $D \times D \times HWC$. In contrast, our GSSA approach only has a constant time complexity $D\times D \times C$, where $D$ denotes the number of bands.

\textbf{Learnable Query.} The fourth notable difference is the introduction of the learnable query (LQ), which is motivated by the fixed patterns of pixel values across different bands. For example, the values of band 100 nm and 200 nm are correlated. Our LQ helps to identify these correlations and the alternative training strategy enables improvements without any extra cost on the number of parameters, inference time, and the flexibility to handle HSIs with different bands.

\section{More Ablation Studies}

To evaluate the effectiveness of the proposed components, we conduct a series of experiments to explore the different design choices for each part of our HSDT architecture. 
Specifically, we compare the proposed blocks, which include GSSA, S3Conv, and SM-FNN, by separately replacing them with existing blocks that share the same functionality, \eg, replacing S3Conv with Conv3D. We use HSDT-M as the base model and evaluate the performance of the different blocks by replacing them one at a time. For blocks that cannot be incorporated into our 3D architectural design of HSDT, such as 2D spectral attention \cite{li2022spatial}, we report the results obtained using their respective models.

\begin{table}[t]
      \setlength{\tabcolsep}{0.2cm}
       \centering
       \small
       \begin{tabular}{l| c| c | c| c}
         \hline
      \textbf{Model}  & \textbf{\#P(Conv)} & \textbf{\#P(Total)} & \textbf{PSNR} & \textbf{SAM} \\
       \hline
       Conv3D & 0.43M & 0.58M & 41.62 & 0.052\\
       Sep3D \cite{dong2019deep}  & 0.37M & 0.53M & 41.44 & 0.054\\
\rowcolor{graycolor} S3Conv-S  & 0.26M & 0.42M & 41.47 & 0.052\\
      \rowcolor{graycolor} S3Conv-Seq  & 0.26M & 0.42M & 41.58 & 0.052\\
     \rowcolor{graycolor}  S3Conv  & 0.36M & 0.52M & \textbf{41.82} & \textbf{0.049} \\
       \hline
      \end{tabular}
      \caption{Comparison of different S3Conv variants against 3D convolution and previous separable convolution. Our S3Conv achieves significant better performance with fewer parameters. Our methods are highlighted as \colorbox{graycolor}{gray}. \#P denotes the model parameters. }
      \label{tab:ablation-s3conv}
\end{table}

\vspace{-3mm}

\paragraph{Spatial-Spectral Separable Convolution.} We evaluate several variants of our S3Conv. The most straightforward variant, S3Conv-S, sets the number of spatial convolutions to 1, while the S3Conv variant that we adopt uses 2. Another variant, S3Conv-Seq, applies spatial and spectral convolutions sequentially instead of in parallel. As shown in Table \ref{tab:ablation-s3conv}, both variants achieve comparable performance with roughly 60\% of the parameters used by Conv3D. Our adopted version achieves a 0.2 dB PSNR gain with only 80\% of the parameters used by Conv3D. Notably, our S3Conv approach significantly outperforms previous HSI separable convolution approaches \cite{dong2019deep}, achieving over 0.4 dB PSNR improvement with even fewer parameters.

\vspace{-3mm}

\paragraph{Guided Spectral Self-Attention.} We compare the proposed GSSA approach with existing spectral fusion techniques, including QRU \cite{wei20203}, GSA \cite{li2022spatial}, MS-MSA \cite{cai2022mask}, MDTA \cite{zamir2022restormer}, and MGSA \cite{chen2022hider}. It is worth noting that although GSA and MS-MSA are named as spectral attention, they are essentially channel attentions derived from MDTA, as discussed earlier. Furthermore, GSA, MS-MSA, and MDTA are all 2D attention approaches that work with 4D data formats instead of the 5D data format used by HSDT. Therefore, we report the results of their models when compared with GSA, MS-MSA, and MDTA. For 3D spectral fusion techniques such as QRU and MGSA, we report the results of models that replace the GSSA of HSDT-M with them. Table \ref{tab:ablation-ssa} presents the results of different attention mechanisms. Our GSSA approach achieves the best results against the other approaches. Notably, our GSSA outperforms previous GSA and MGSA approaches (which are also designed for HSI denoising) by a large margin, demonstrating the effectiveness of our designs.

\begin{table}[ht]
   \setlength{\tabcolsep}{0.3cm} 
       \centering
       \small
       \begin{tabular}{l | l | l | l}
         \hline
      \textbf{Model}  & \textbf{Params} & \textbf{PSNR} & \textbf{SAM} \\
       \hline 
     QRU \cite{wei20203}  & 0.57M & 41.31 & 0.064\\
     GSA \cite{li2022spatial} \& MS-MSA \cite{cai2022mask} & 4.14M & 41.41 & 0.052\\
     MDTA \cite{zamir2022restormer} & 26.2M & 41.03 & 0.062\\
     MGSA \cite{chen2022hider} & 0.50M & 39.74 & 0.102\\
 \rowcolor{graycolor} GSSA & 0.52M & \textbf{41.82} & \textbf{0.049}\\
       \hline
      \end{tabular}
 	\caption{Results of our GSSA in comparison with other attention blocks. Our GSSA achieves a prominent improvement against QRU by over 0.5 PSNR improvement, while previous HSI denoising transformer with GSA only outperforms QRU by only 0.1 PSNR. }
      \label{tab:ablation-ssa}
\end{table}

\vspace{-3mm}
\paragraph{Self-Modulated Feed-Forward Network.} The proposed {SM-Branch} can be used without additional conventional FFN. As shown in \tabref{tab:ablation-ffn}, the sole use of {SM-Branch} also outperforms the conventional FFN, and the combination of them both yields the best results with very few extra parameters. The {GDFN} \cite{zamir2022restormer} developed for RGB restoration performs poorly and might be unsuitable for our model.

  \begin{table}[h]
    \setlength{\tabcolsep}{0.5cm}
       \centering
       \small
       \begin{tabular}{l | l | l | l}
         \hline
   \textbf{Model}  & \textbf{Params} & \textbf{PSNR} & \textbf{SAM} \\
       \hline
       FFN & 0.49M & 41.67 & 0.050\\
       GDFN \cite{zamir2022restormer} & 0.49M &37.38 & 0.094\\
   \rowcolor{graycolor}      SM-Branch  & 0.45M &41.74 & 0.051\\
    \rowcolor{graycolor}     SM-FFN  & 0.52M & \textbf{41.82} & \textbf{0.049} \\
       \hline
      \end{tabular}
      \caption{Comparison of the existing FFN with our SM-FFN and SM-Branch.  }
      \label{tab:ablation-ffn}
   \end{table}

\section{More Discussions} \label{sec:discussion}

\begin{figure}[h]
  \centering
  \includegraphics[width=1\linewidth]{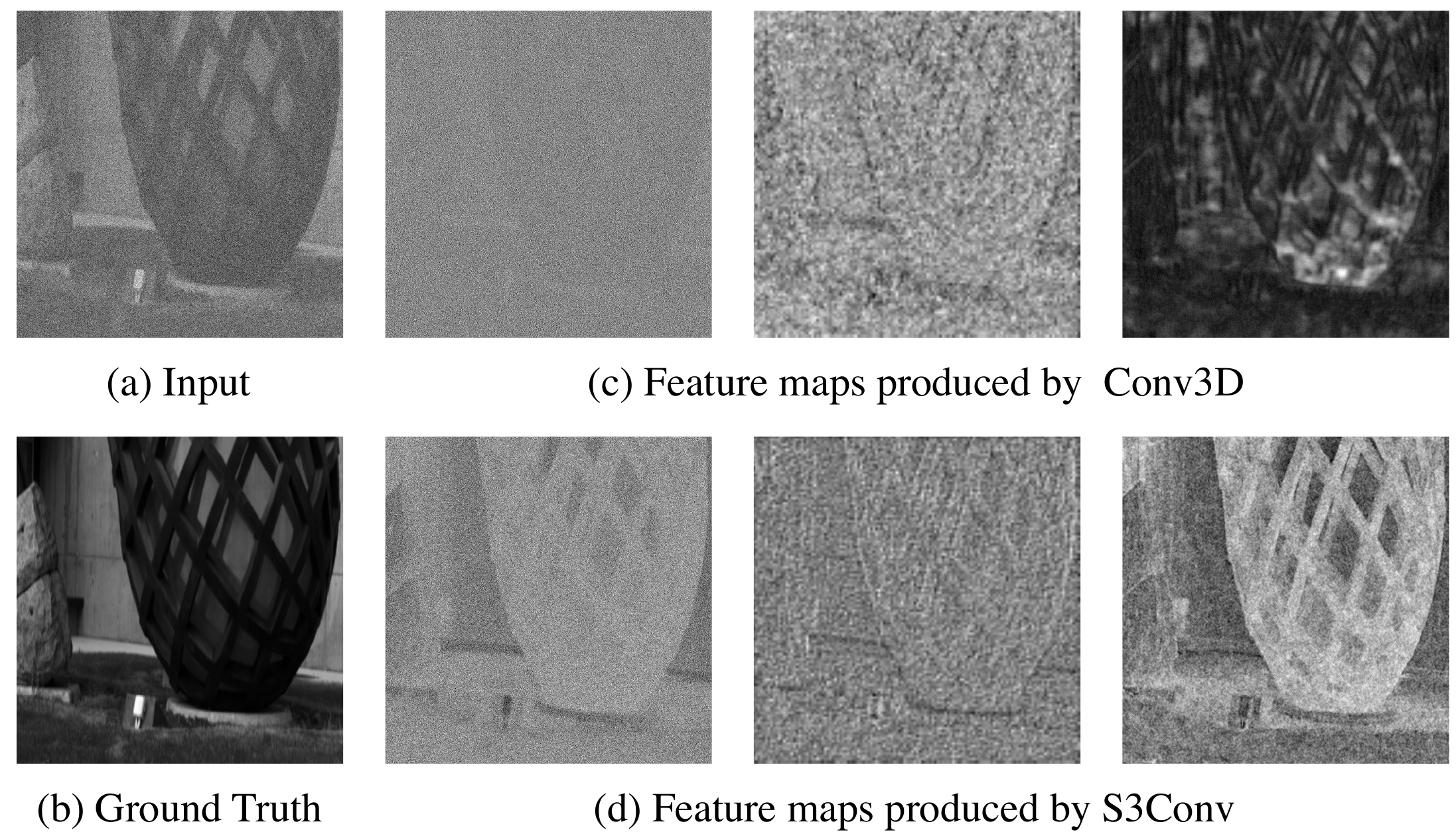}
  \caption{Comparison of the feature maps extracted by conventional 3D  convolution and our S3Conv. }
  \label{fig:s3conv}
  \vspace{-3mm}
\end{figure}

\paragraph{Visualization of S3Conv.} To demonstrate the effectiveness of our S3Conv. We provide a comparison of the features map between S3Conv and conventional 3D convolution. As shown in \figref{fig:s3conv}, our S3Conv extracts more spatial meaningful features.

\vspace{-3mm}
\paragraph{Analysis of SM-FFN.} The proposed SM-FFN is designed for strengthening the features with higher activation via a self-modulation operation. The improvement provided by SM-FFN could be intuitively explained by the emphasis on more informative regions that typically have higher activation. In the following, we provide some possible relations between our SM-FFN and the SiLU \cite{elfwing2018sigmoid} activation, which might further imply why our SM-FFN works better. Specifically, The SiLU activation is,
\begin{equation}
	y = x\odot \operatorname{sigmoid}(x),
\end{equation}
where $x$ and $y$ are the input and output feature maps. It can be observed that SiLU could be treated as a kind of self-modulation where the modulation weight is computed from the input itself. However, such homogeneous self-modulation might be limited in expressive abilities. Instead, our SM-FFN employs a heterogeneous self-modulation, 
\begin{equation}
	y = \operatorname{Linear_1}(x)\odot \operatorname{sigmoid}(\operatorname{Linear_2}(x)),
\end{equation}
where we adopt two extra linear projections to project input x into two different spaces. This removes the restriction of SiLU where the input $x$ should simultaneously play two roles of features and modulation weight. Thus, our SM-FFN can obtain the advantages of SiLU, \eg, training stability and implicit regularization while maintaining more representation capability. Consequently, it leads to better performance than conventional FFN.

\begin{table*}[t]
\begin{center}
   \begin{subtable}[h]{0.4\textwidth}
      \setlength{\tabcolsep}{0.35cm}
       \centering
       \small
      \begin{tabular}{|l|cc|}
\hline
\rowcolor{graycolor} Method          & PSNR       & SSIM             \\ \hline
2DTV            & 25.26      & 0.863            \\
3DTV            & 28.46      & 0.910            \\
DeSCI \cite{Liu19DeSCI}          & 26.62      & 0.912             \\
SCI-TV-FFDNet \cite{qiu2021effective}  & 29.35      & 0.925           \\ 
DPHSIR \cite{lai2022deep}       & 30.56     & 0.945           \\
\hline

PnP-HSDT (ours) & \textbf{31.64}    & \textbf{0.948}                             \\ \hline
\end{tabular}
      \caption{Results on the task of compressive sensing.}
      \label{tab:cs}
   \end{subtable}
   \hfill
   \begin{subtable}[h]{0.55\textwidth}
   \setlength{\tabcolsep}{0.3cm}
       \centering
       \small
\begin{tabular}{|l|cccc|}
\hline
\rowcolor{graycolor} & \multicolumn{2}{c}{2x}  &  \multicolumn{2}{c|}{4x}\\
\rowcolor{graycolor} Method          & PSNR       & SSIM     & PSNR & SSIM        \\ \hline
Bicubic                     & 35.13      & 0.9575 &35.12  &0.954         \\
SSPSR \cite{jiang2020learning}          & 47.55      & 0.995   &39.19 &0.979          \\
Bi-3DQRNN \cite{fu2021biqrnn3d}                   & 42.53      & 0.989    &39.56 &0.979        \\
DPHSIR \cite{lai2022deep}       & 48.75     & 0.996   &40.95 &0.980        \\
\hline
PnP-HSDT (ours) & \textbf{49.76}    & \textbf{0.996}   & \textbf{41.56}    & \textbf{0.982}                          \\ \hline
\end{tabular}
      \caption{Results on the task of super-resolution.}
      \label{tab:sr}
   \end{subtable}
\end{center}
\vspace{-5mm}
\caption{Experimental results of our PnP extension on the task of compressive sensing and super-resolution.}
\end{table*}

\section{Extension as Plug-and-Play Prior} \label{sec:ext}
Considering the superior performance of our method on the Gaussian denoising task, 
we demonstrate that HSDT can be used a plug-and-play (PnP) prior \cite{chan2016plug} to solve general HSI restoration tasks with proximal algorithms, \eg, ADMM and HQS.

\vspace{-3mm}
\paragraph{Experimental Setup.} We adopt PnP-ADMM \cite{lai2022deep} to extend our method to the tasks of compressive sensing, and super-resolution. To meet the requirements of PnP algorithms, \ie, Gaussian denoiser for continuous noise strengths, we retrain our model, \ie, HSDT-M, with an additional noise level map \cite{zhang2020plug} on simulated Gaussian noise ranged from 0 to 70. We run 40 iterations for compressive sensing and 24 iterations for super-resolution. The hyperparameters of the algorithms are manually tuned to achieve the best performance.

\vspace{-3mm}
\paragraph{Compressive Sensing.} We conduct the simulated experiments on CASSI \cite{wagadarikar2009video} system. Following \cite{qiu2021effective}, the shifting random binary mask \cite{llull2013coded} is used in our simulation. We provide the results on CAVE \verb+Toy+, which is obtained from \cite{Liu19DeSCI}. We compare several recent methods, including DPHSIR \cite{lai2022deep}, SCI-TV-FFDNet \cite{qiu2021effective}, DeSCI \cite{Liu19DeSCI}, and traditional methods, \ie, 2DTV and 3DTV. The quantitative results are shown in Tab.~\textcolor{linkcolor}{1}\subref{tab:cs}. It can be seen that our method obtains the best performance with over 1 dB improvement on PSNR. Specifically, the improvement is purely obtained through the superior denoising ability of our model, which means our model can also be integrated into other more advanced PnP methods for further improvement, \eg, \cite{qiu2021effective}.

\vspace{-3mm}
\paragraph{Super-Resolution.} We also provide results on the task of HSI super-resolution. Following \cite{lai2022deep}, we first blur the high-resolution HSI via an $8\times 8$ Gaussian blur kernel with $\sigma=3$, and then downsample the image to obtain the low-resolution HSI. We provide the results on ICVL with a scale factor of 2 and 4. The competing methods include several recently developed methods, \eg, SSPSR \cite{jiang2020learning}, Bi3DQRNN \cite{fu2021biqrnn3d}, and DPHSIR \cite{lai2022deep} . As shown in Tab.~\textcolor{linkcolor}{1}\subref{tab:sr}, our method achieves the best performance. In particular, our method only needs the pretrained Gaussian denoising model, which is the same as \cite{lai2022deep}. The improvement against \cite{lai2022deep} comes from the better PnP denoising prior, which further demonstrates the stronger denoising ability of our method.

\section{More Implementation Details}  \label{sec:impl}

\begin{table*}[t]
\begin{center}
   \begin{subtable}[h]{0.56\textwidth}
      \setlength{\tabcolsep}{0.1cm}
       \centering
       \small
      \begin{tabular}{|l|cccccc|}
\hline
\rowcolor{graycolor} Stage 1 & \multicolumn{6}{c|}{Gaussian Noise $\sigma=50$}                \\ \hline
Epoch & 0 - 20  & 20 - 30 &        &         &         &       \\
LR & $1\times 10^{-3}$     &  $1\times 10^{-4}$     &        &         &         &       \\ \hline
\rowcolor{graycolor}Stage 2  & \multicolumn{6}{c|}{Gaussian Noise $\sigma=10,30,50,70$}                \\ \hline
Epoch & 30 - 45 & 45 - 55 & 55 - 60  & 60 - 65   & 65 - 75   & 75 - 80 \\
LR & $1\times 10^{-3}$     &  $1\times 10^{-4}$     &  $5\times 10^{-5}$      &         $1\times 10^{-5}$ & $5\times 10^{-6}$        & $1\times 10^{-6}$      \\ \hline
\rowcolor{graycolor}Stage 3 & \multicolumn{6}{c|}{Complex Noise}                 \\ \hline
Epoch & 80 - 90 & 90 - 95 & 95 - 100 & 100 - 105 & 105 - 110 &       \\
LR &   $1\times 10^{-3}$   & $5\times 10^{-4}$      & $1\times 10^{-4}$          & $5\times 10^{-5}$        & $1\times 10^{-5}$   &   \\ \hline
\end{tabular}
      \caption{Our multi-step learning rate scheduler.}
      \label{tab:lr}
   \end{subtable}
   \hfill
   \begin{subtable}[h]{0.4\textwidth}
   \setlength{\tabcolsep}{0.21cm}
       \centering
       \small
       \begin{tabular}{|>{\columncolor{graycolor}}l|l|}
\hline
System     &  Ubuntu 20.04.1 LTS \\ \hline
GPU        &  Nvidia GeForce RTX 3090 \\ \hline
CPU        &  Intel(R) Core(TM) i9-10850K CPU \\ \hline
Framework  &  PyTorch 1.7.1 \\ \hline
Driver  &  Cuda 11.2 \\ \hline
Software   &  Matlab 2020 \\ \hline
Dataset    &  ICVL  \\ \hline
Image Size &  $512 \times 512$ \\ \hline
Repeat times &  10 \\ \hline
\end{tabular}
      \caption{System configuration for the speed test.}
      \label{tab:system}
   \end{subtable}
\end{center}
\vspace{-4mm}
\caption{More implementation details. (a) We adopt a multi-stage training strategy with the learning warmup setup for the first epoch. (b) We provide the system configuration as the results of the speed test are strongly correlated with the configuration. }
\end{table*}

\paragraph{Setup of the Learning Rate.} In this part, we provide more details about the multi-step learning rate scheduler that we used for training our simulated Gaussian and complex denoising models. Specifically, we use a multi-stage training strategy to train the models for Gaussian noise and complex noise. The learning rate is set up as shown in Tab.~\textcolor{linkcolor}{3}\subref{tab:lr}. We use learning rate warmup to gradually increase the learning rate from 0 to $1\times 10^{-3}$ for the first epoch of the second stage.

\vspace{-4mm}
\paragraph{Details of the Simulated Complex Noise.} 

We follow \cite{wei20203} for constructing simulated complex noise. In details, we consider the non-independent and non-identically distributed (non-i.i.d) Gaussian noise, stripe noise, deadline noise, impulse noise, and the combination of the aforementioned noise (denoted as mixture noise). The details about these five cases of noise are listed as follows,
\begin{itemize}[noitemsep, leftmargin=*]
	\item \textbf{Non-i.i.d noise}. The non-independent and non-identically distributed Gaussian is added to every pixel of each HSI. The noise strength is randomly selected from 10, 30, 50, and 70.
	\item \textbf{Stripe noise}. Stripe noise (5\% to 15\% percentages of columns) is added to randomly selected one-third of bands. Non-i.i.d. Gaussian noise is added to All bands. 
	\item \textbf{Deadline noise}. Deadline noise is added to randomly selected one-third of bands. Non-i.i.d. Gaussian noise is added to All bands.  
	\item \textbf{Impulse noise}. Impulse noise with intensity ranging from 10\% to 70\% is added to randomly selected one-third of bands. Non-i.i.d. Gaussian noise is added to All bands. 
	\item \textbf{Mixture noise}. Each band is randomly corrupted by at least one kind of noise mentioned above.
\end{itemize}

\vspace{-4mm}
\paragraph{System Configuration.} In the main paper, we compare the running time of different methods. All the comparisons are performed with an Nvidia GeForce RTX 3090, and an Intel(R) Core(TM) i9-10850K CPU @ 3.60GHz on Ubuntu 20.04.1 LTS. All the CNN-based methods are implemented and tested with PyTorch 1.7.1. All the optimization-based methods are implemented and tested with Matlab. We test the running time on ICVL with an image size of $512 \times 512$ by repeating the test 10 times and averaging the results.

\section{Future work.}   \label{sec:future}

In this work, we propose a transformer architecture, \ie, HSDT for hyperspectral image denoising. We introduce several effective and generalizable components to better explore the spatial-spectral and global spectral correlations of HSI. Specifically, it is worthwhile to explore the applications of the proposed S3Conv and HSDT for more network architectures and tasks. Furthermore, our learnable queries could also be extended to condition on some external information for more explicit guidance. For example, we might be able to inject the Gaussian noise strength into the network with learnable queries, through an embedding layer. This is helpful for a PnP Gaussian denoiser, where the noise strength is known.

\section{Broader Impacts}  \label{sec:impact}
Our work has no ethical issues or broader impacts.